\documentclass[10pt,journal,compsoc]{IEEEtran}
\ifCLASSOPTIONcompsoc
  \usepackage[nocompress]{cite}
\else
  \usepackage{cite}
\fi

\ifCLASSINFOpdf
\else
\fi

\usepackage{graphicx}
\usepackage{algorithm}
\usepackage{algpseudocode}
\usepackage{amsmath}
\usepackage{amssymb}
\usepackage{amsfonts}
\usepackage{multicol}
\usepackage{multirow}
\usepackage{color, graphics}
\usepackage{epsfig}
\usepackage{epstopdf}
\usepackage{balance}
\usepackage{verbatim}

\newtheorem{theorem}{\bf Theorem}
\newtheorem{prop}{\bf Proposition}

\newtheorem{remark}{\bf Remark}

\def\blue{\textcolor{blue}}

\def\blue{\textcolor{black}}

\def\ie{\emph{i.e.}}
\def\eg{\emph{e.g.}}
\def\yifan{\textcolor{black}}

\usepackage{mathrsfs}
\usepackage{epsfig}

\usepackage{caption}

\begin{document}

\title{Cost-Sensitive Portfolio Selection via Deep Reinforcement Learning}

%


\author{Yifan Zhang, Peilin Zhao, Qingyao Wu, Bin Li, Junzhou Huang, and Mingkui Tan 
\IEEEcompsocitemizethanks{
\IEEEcompsocthanksitem Y. Zhang, Q. Wu and M. Tan are with South China University of Technology and Guangzhou  Laboratory, China. E-mail: sezyifan@mail.scut.edu.cn; $\{$qyw, mingkuitan$\}$@scut.edu.cn. 
\IEEEcompsocthanksitem P. Zhao and J. Huang are with Tencent AI Lab, China. Email: peilinzhao@hotmail.com; joehhuang@tencent.com.
\IEEEcompsocthanksitem B. Li is with Wuhan University, China. E-mail: binli.whu@whu.edu.cn. 
\IEEEcompsocthanksitem P. Zhao, Q. Wu are co-first authors. M. Tan is the corresponding author.}
}

\markboth{IEEE TRANSACTIONS ON KNOWLEDGE AND DATA ENGINEERING}%
{Zhang \MakeLowercase{\textit{et al.}}}

\IEEEtitleabstractindextext{
\begin{abstract}
Portfolio Selection is an important real-world financial task and has attracted extensive attention in artificial intelligence communities. This task, however, has two main difficulties: (i) the non-stationary price series and complex asset correlations make the learning of feature representation very hard; (ii) the practicality principle in financial markets requires controlling both transaction and risk costs. Most existing methods adopt handcraft features and/or consider no constraints for the costs, which may make them perform unsatisfactorily and fail to control both costs in practice. In this paper, we propose a cost-sensitive portfolio selection method with deep reinforcement learning. Specifically, a novel two-stream portfolio policy network is devised to extract both price series patterns and asset correlations, while a new cost-sensitive reward function is developed to maximize the accumulated return and constrain both costs via reinforcement learning. We theoretically analyze the near-optimality of the proposed reward, which shows that the growth rate of the policy regarding this reward function can approach the theoretical optimum. We also empirically evaluate the proposed method on real-world datasets. Promising results demonstrate the effectiveness and superiority of the proposed method in terms of profitability, cost-sensitivity and representation abilities.
\end{abstract}

}

\maketitle

\IEEEdisplaynontitleabstractindextext

\IEEEpeerreviewmaketitle

\IEEEraisesectionheading{\section{Introduction}\label{Introduction}}

\IEEEPARstart{P}{ortfolio}  Selection~\cite{markowitz1952portfolio} aims at dynamically allocating the wealth among a set of assets to maximize the long-term return.
This task, however, is difficult for many individual investors, since even an expert has to spend a lot of time and efforts to deal with each asset with professional knowledge.
Recently, many intelligent portfolio selection methods have been proposed and have shown remarkable improvement in performance \cite{das2014online,huang2015semi,zhao2016,cao2012,shen2015transaction,jiang2017deep,ding2018investor}.
However, these methods can be limited in practice, due to two main challenges brought by \yifan{the} complex nature of portfolio selection as follows.

One of the key challenges in portfolio selection is how to represent the non-stationary price series, since the asset price sequences often contain a large number of noises, jumps and oscillations.
Most existing methods use handcraft features, such as moving average~\cite{li2012line} and stochastic technical indicators~\cite{neely2014forecasting}, which, however, perform unsatisfactorily because of poor representation abilities~\cite{deng2017deep}.
In recent years, deep neural networks (DNNs) have shown strong representation abilities in modeling sequence data~\cite{sutskever2014sequence} and often lead to better performance~\cite{lecun2015deep}.
However, it is non-trivial for existing DNNs to directly extract price sequential patterns and asset correlations simultaneously.
Nevertheless, both kinds of information significantly affect the decision-making for portfolio selection.
More critically, the dynamic nature of portfolio selection and the lack of well-labeled data make DNNs hard to train.

Another key challenge of portfolio selection is how to control costs in decision-making, since the transaction cost and the risk cost highly affect the practicality of algorithms.
The transaction cost~(\eg, tax and commission) is common in decision-making~\cite{neuneier1996optimal,li2017transaction}. Ignoring this cost may lead to aggressive trading~\cite{das2013online} and bring biases into the estimation of returns~\cite{ormos2013performance}.
The risk cost is incurred by the fluctuation of returns and is an important concern in financial investment~\cite{neuneier1999risk}.
\yifan{Neglecting this cost may lead to a disastrous consequence in practice~\cite{heger1994consideration}.
Most existing methods consider either one of them but do not constrain both costs simultaneously, which may limit their practical performance.}

In this paper, considering the challenges of portfolio selection and its dynamic nature, we formulate portfolio selection as a Markov Decision Process (MDP), and \yifan{propose a cost-sensitive portfolio policy network (PPN)} to address it via reinforcement learning.
Our main contributions are summarized as follows.

  $\bullet$  To extract meaningful features, we devise a novel two-stream network architecture to capture both price sequential information and asset correlation information. With such information, PPN makes more profitable decisions.

  $\bullet$  To control both transaction and risk costs, we develop a new cost-sensitive reward function.  By exploiting reinforcement learning to optimize this reward function, the proposed PPN is able to maximize the accumulated return while controlling both costs.

  $\bullet$  We theoretically analyze the near-optimality of the proposed reward. That is, the wealth growth rate regarding this reward function can be close to the theoretical optimum.

  $\bullet$  Extensive experiments on real-world datasets demonstrate the effectiveness and superiority of  the proposed method  in terms of profitability, cost-sensitivity and representation abilities.

\section{Related Work}

Following Kelly investment principle~\cite{kelly1956new}, many kinds of portfolio selection methods have been  proposed, including online learning and reinforcement learning based methods.

Online learning based methods maximize the expected log-return with sequential decision-making.
The pioneering studies include Constant Rebalanced Portfolios (CRP)~\cite{cover1986empirical,cover1991universal},  Universal Portfolios (UP)~\cite{cover1991universal}, Exponential Gradient (EG)~\cite{helmbold1998line}, Anti-correlation (Anticor)~\cite{borodin2004can} and Online Netwon Step (ONS)~\cite{agarwal2006algorithms}.
Recently, several methods exploit the mean reversion property to select the portfolio, e.g., Confidence Weighted Mean Reversion (CWMR)~\cite{li2011confidence}, Passive Aggressive Mean Reversion (PAMR)~\cite{li2012pamr}, Online Moving Average Reversion (OLMAR)~\cite{li2012line}, Robust Median Reversion (RMR)~\cite{huang2013robust} and Weighted Moving Average Mean Reversion (WMAMR)~\cite{gao2013weighted}. \blue{In addition, the work~\cite{Shen2019}  proposes an ensemble learning method for Kelly growth optimal portfolio.} 

However, all the above methods ignore the learning of sequential features and only use some handcraft features, such as moving average and stochastic technical indicators. As a result, they may perform unsatisfactorily due to poor representation abilities~\cite{deng2017deep}. More critically, many of the above methods assume no transaction cost. Such a cost will bring biases into the estimation of accumulative returns~\cite{ormos2013performance}, and thus affects the practical performance of these methods. In contrast, our proposed method not only learns good feature representation based on the proposed two-stream architecture, but is also sensitive to both costs.

On the other hand, reinforcement learning based methods use reinforcement learning algorithms to optimize specific utility functions and make comprehensive policies~\cite{moody1998reinforcement,moody1998performance,neuneier1999risk,moody2001learning,neuneier1996optimal,neuneier1998enhancing}.
However, all these methods ignore the feature representation on portfolios.
Very recently, some studies apply deep reinforcement learning to portfolio selection, where they use deep neural networks~\cite{chen2020intelligent,guo2019nat,zeng2020dense,zeng2019graph,ref_zhang2019miccai, ref_zhang2019miccai} to extract features~\cite{jiang2017deep,guo2018robust}.
Specifically, the state-of-the-art one is the ensemble of identical independent evaluations (EIIE)~\cite{jiang2017deep}.
However, both methods~\cite{jiang2017deep,guo2018robust} ignore the asset correlation and do not control costs during optimization, leading to limited representation abilities and performance. 
In contrast, our method can control both kinds of costs relying on the new proposed cost-sensitive reward.

Beyond that, there are also some theoretical studies on the optimal portfolio.
To be specific, a theoretical optimal policy can be obtained by maximizing the expected log-return~\cite{algoet1988asymptotic}.
Based on this, a mean-variance portfolio selection is studied~\cite{ottucsak2007asymptotic}.
However, both studies assume no transaction cost, making them less practical.
When considering the transaction cost, a theoretical optimal strategy can be achieved by optimizing the expected rebalanced log-return~\cite{gyorfi2008growth}.
This work, however, ignores the risk cost.
Instead, in this paper, we provide theoretical analyses for the proposed reward in the presence of both  costs.

\section{Problem Settings}

Consider a portfolio selection task over a financial market during $n$ periods with $m\small{+}1$ assets, including one cash asset and $m$ risk assets.
On the $t$-th period, we denote the prices of all assets as $p_t \small{\in} \mathbb{R}^{(m+1)\times d}_{+}$, where each row $p_{t,i}\small{\in} \mathbb{R}^{d}_{+}$ indicates the feature of asset $i$, and $d$ denotes the number of prices.
Specifically, we set $d\small{=}4$ in this paper.
That is, we consider four kinds of prices, namely the opening, highest, lowest and closing prices.
One can generalize it to more prices to obtain more information.
The price series is represented by $P_{t}\small{=} \{p_{t-k},..,p_{t-1}\}$, where $k$ is the length of the price series.

The price change on the $t$-th period is specified by a \emph{price relative vector} $x_t \small{=} \frac{p_{t}^{c}}{p_{t-1}^{c}}\in \mathbb{R}^{m+1}_{+}$, where $p_{t}^{c}$ is the closing price of assets.
Typically, $x_{t,0}$ represents the price change of the cash asset.
\yifan{Assuming there is no inflation or deflation, the cash is risk-free with invariant price, \ie, $\{\forall t|x_{t,0}\small{=}1\}$, and it has little influence on the learning process.
We thus exclude the cash asset in the input, \ie, $P_t \small{\in} \mathbb{R}^{m \times k \times 4}$.
When making decisions, the investment decision is specified by a \emph{portfolio vector} $a_t\small{=}[a_{t,0},a_{t,1},a_{t,2},\ldots,a_{t,m}] \in\mathbb{R}^{m+1}$, where $a_{t,i} \small{\geq} 0$ is the proportion of asset $i$, and $\sum_{i=0}^{m} a_{t,i}\small{=}1$.}
\yifan{Here, the portfolio decision contains the proportion of all assets, including the cash $a_{t,0}$.}
We initialize the \emph{portfolio vector} as $a_0\small{=}[1,0,..,0]$ and initialize the gross wealth as $S_0\small{=}1$.
After $n$ periods, the accumulated wealth, if ignoring the transaction cost $c_t$,  is $S_n\small{=}S_0 \prod_{t=1}^n a_t^{\top}x_t$;  otherwise, $S_n\small{=}S_0 \prod_{t=1}^n a_t^{\top}x_t(1-c_t)$.

\yifan{There are two general assumptions~\cite{li2012line,Vajda2006Analysis} in this task:} (i) perfect liquidity: each investment can be carried out immediately; (ii) zero-market-impact: the investment by the agent has no influence on the financial market, \ie, the environment. 

\subsection{Markov Decision Process for Portfolio Selection}
\begin{figure}[!htp]
    \vspace{0.1in}
  \centerline{\includegraphics[width=8cm]{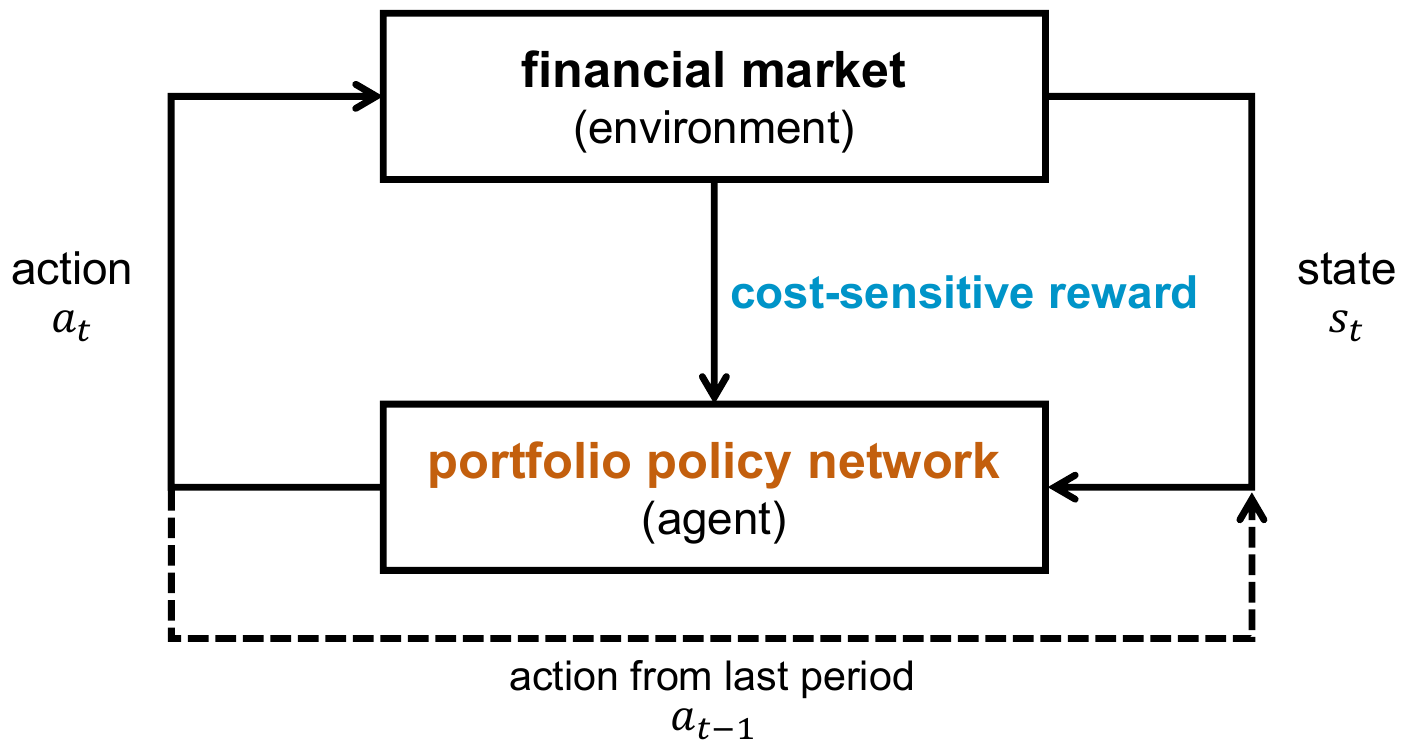}}
  \caption{\yifan{Markov decision process for portfolio selection (Better viewed in color)}}
  \label{fig1}
\end{figure}

We formulate the investment process as a generalized Markov Decision Process by ($\mathcal{S},\mathcal{A},\mathcal{T},\mathcal{R}$).
Specifically, as shown in Fig.~\ref{fig1}, on the $t$-th period, the agent observes a state $s_t \small{=} P_t\normalsize{\in}\mathcal{S}$, and takes an action $a_t\small{=} \pi(s_t,a_{t-1})\normalsize{\in}\mathcal{A}$, which determines the reward $r_t\small{=}a_t^{\top}x_t \normalsize{\in}\mathcal{R}$, while the next state is a stochastic transition $s_{t+1}\normalsize{\sim}\mathcal{T}(s_t)$.
\yifan{Specifically}, $\pi(s_t,a_{t-1})$ is a \emph{portfolio policy}, where $a_{t-1}$ is the action of last period.
\yifan{When considering the transaction cost,  the reward will be adjusted as $r_t^c \small{:=} r_t\small{*}(1\small{-}c_t)$, where $c_t$ is the proportion of transaction costs.
In Fig.~\ref{fig1}, portfolio policy network serves as an agent which aims at maximizing the accumulated return while controlling both the transaction and risk costs.}

\begin{remark}
When trading volumes in the financial market are high enough, both general assumptions are near to reality.  Moreover, the assumption (ii) indicates that the action $\mathcal{A}$ will not affect the state transaction $\mathcal{P}$. That is, the state transaction only depends on the environment.
\end{remark}

\begin{figure*} [t]
  \vspace{0.1in}
  \centerline{\includegraphics[width=18cm]{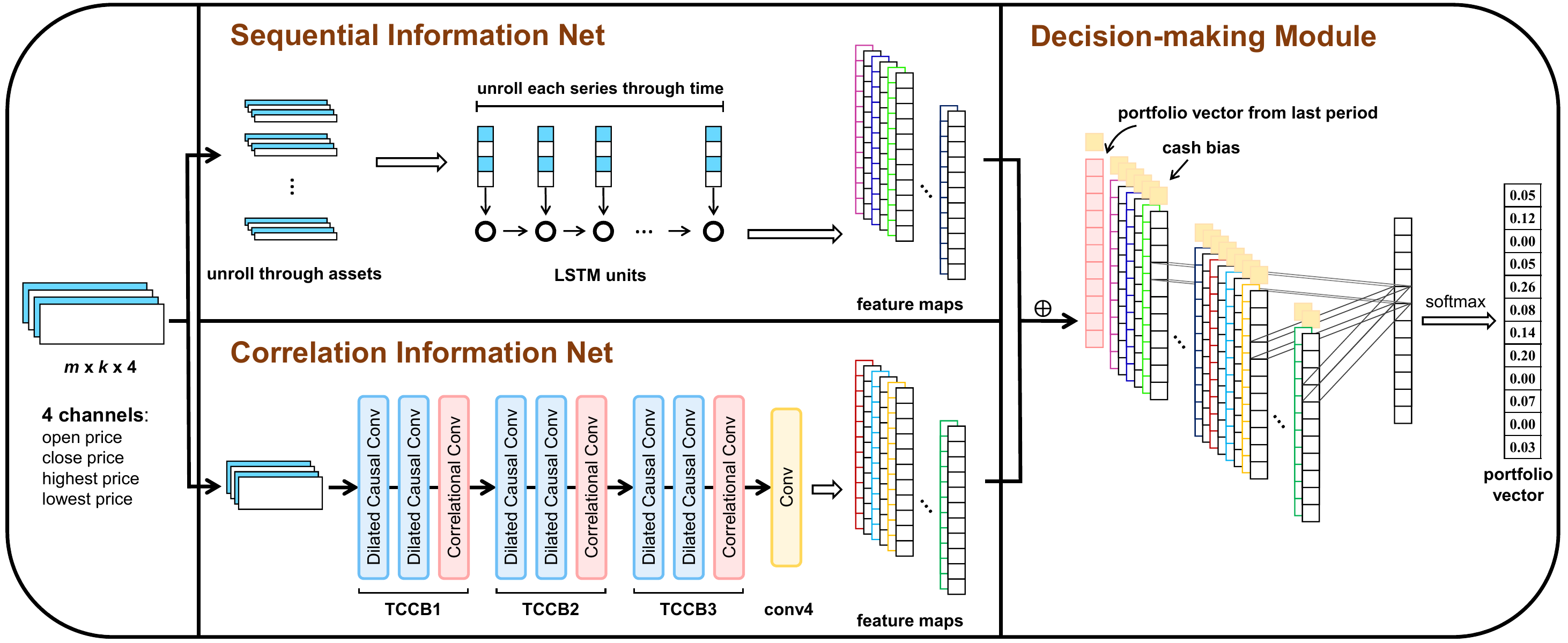}}
  \caption{The scheme of the proposed portfolio policy network, where Correlation Information Net consists of three temporal correlational convolution blocks and $\oplus$ denotes the concatenation operation. More detailed architecture information can be found in Section~\ref{implementation} (Better viewed in color)}
  \label{fig2}
\end{figure*}

\section{Portfolio Policy Network}

\subsection{General Architecture}
In practice, both the price sequential pattern and the asset correlation are significant for the decision-making in portfolio selection tasks.
Specifically, the price sequential pattern reflects the price changes of each asset; while the asset correlation reveals the macro market trend and the relationship among assets.
Therefore, it is necessary to capture both types of information in the learning process.
To this end, we develop a two-stream architecture for portfolio policy network (PPN) to extract portfolio features.
As shown in Fig.~\ref{fig2}, PPN consists of three major components, namely the sequential information net which is to extract price sequential patterns, the correlation information net which is to extract asset correlations, and the decision-making module. Specifically, we will detail these components in the following subsections.

\subsection{Sequential Information Net}

It is non-trivial to extract the price sequential pattern of portfolio series due to the non-stationary property of asset prices.
To solve this issue, we propose a sequential information net, based on LSTM~\cite{hochreiter1997long}, to extract the sequential pattern of portfolios.
This is inspired by the strong ability of LSTM in modeling non-stationary and noisy sequential data~\cite{giles2001noisy}.
Concretely, as shown in Fig.~\ref{fig2}~(top), the sequential information net processes each asset separately, and concatenates the feature of each asset along the height dimension as a whole feature map.
We empirically show that the sequential information net is able to extract good sequential features and helps to gain more profits when only considering the price sequential information (See results in Section~\ref{representation62}).

\subsection{Correlation Information Net}

Although recurrent networks can model the price sequential information, they can hardly extract asset correlations, since they process the price series of each asset separately.
Instead, we propose a correlation information net to capture the asset correlation information based on fully convolution operations~\cite{lea2017temporal,long2015fully,van2016wavenet}.
Specifically, we devise a new temporal correlational convolution block (TCCB) and use it to construct the correlation information net, as shown in Fig.~\ref{fig2}~(bottom).

\begin{figure}[t]
  \centerline{\includegraphics[width=8.8cm]{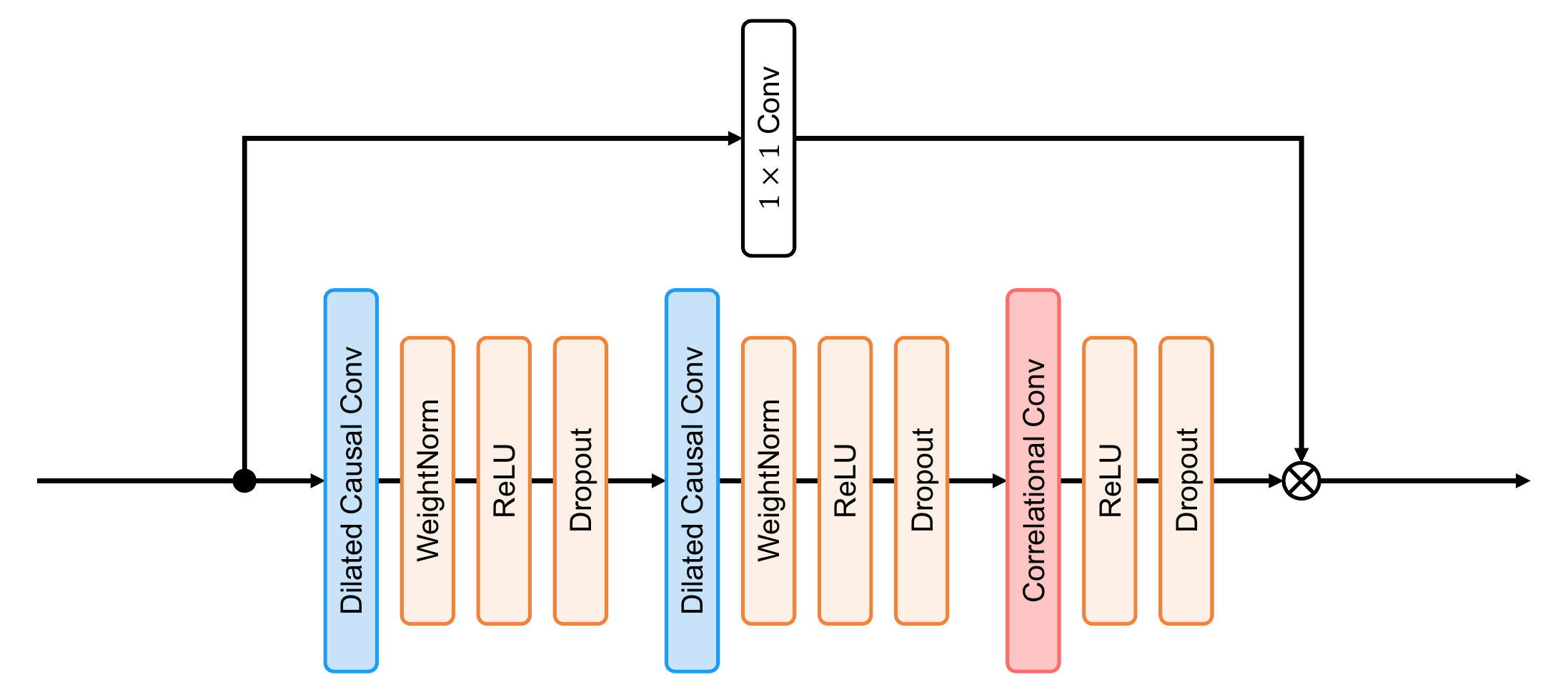}} 
 \caption{Illustration of the temporal correlational convolution block, where $\otimes$ denotes the ReLU activation operation (Better viewed in color)}
 \label{fig3} 
\end{figure}

The proposed TCCB is motivated by the complex nature of portfolio selection. To be specific, we need to extract the asset correlation and model the price series simultaneously.
To this end, we exploit the dilated causal convolution operation to model the portfolio time-series variation, and devise a new correlational convolution operation to capture the asset correlation information.
To make it clear, we summarize the detailed structure of TCCB in Fig.~\ref{fig3}, and describe its two main components as follows.

\subsubsection{Dilated causal convolutions}
\begin{figure}[t]
 \vspace{0.1in}
 \begin{minipage}{0.49\linewidth}
  \centerline{\includegraphics[width=4.3cm]{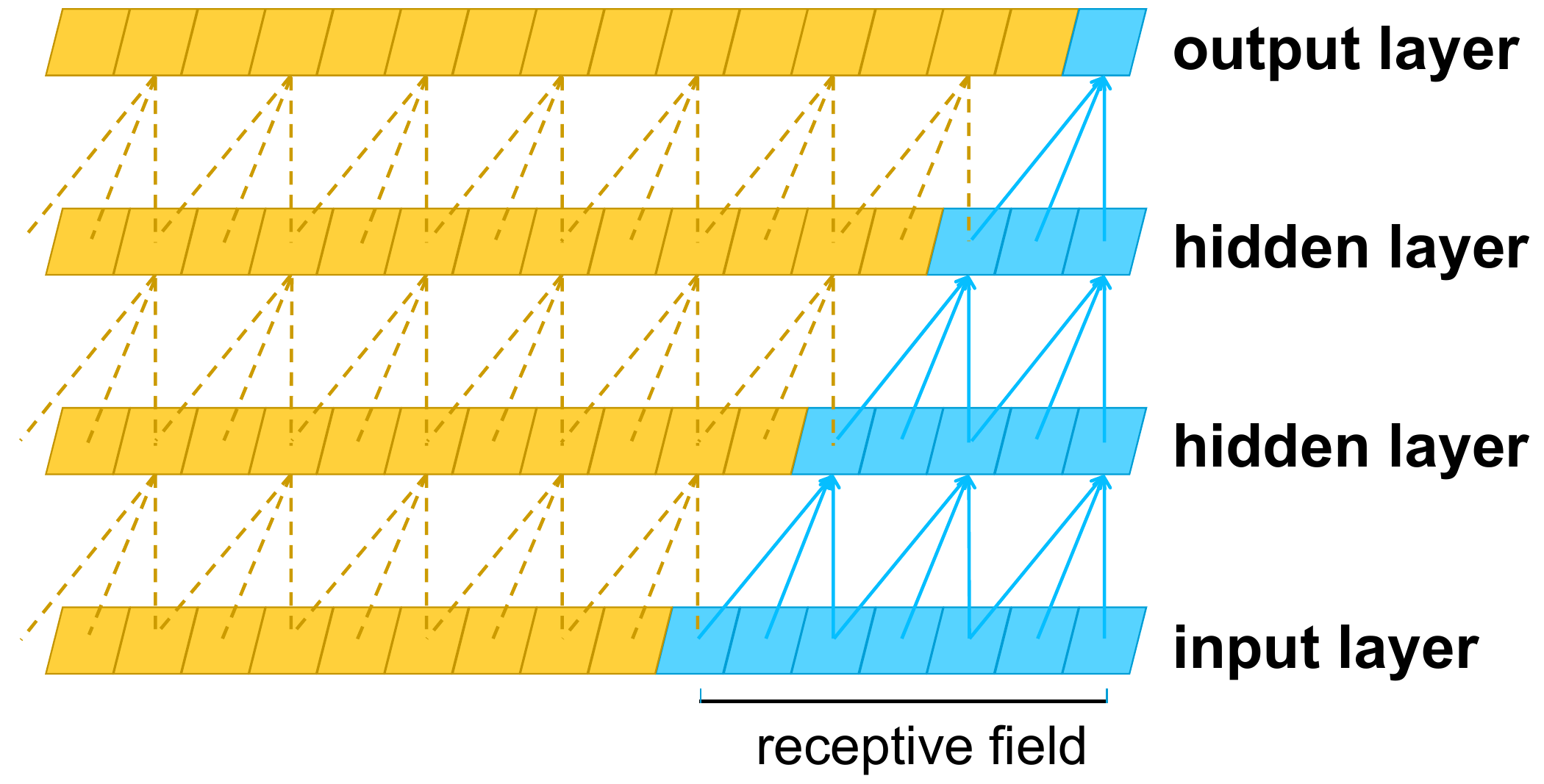}}
  \centerline{(a) Causal convolutions}

 \end{minipage}
 \hfill
 \begin{minipage}{0.49\linewidth}
  \centerline{\includegraphics[width=4.3cm]{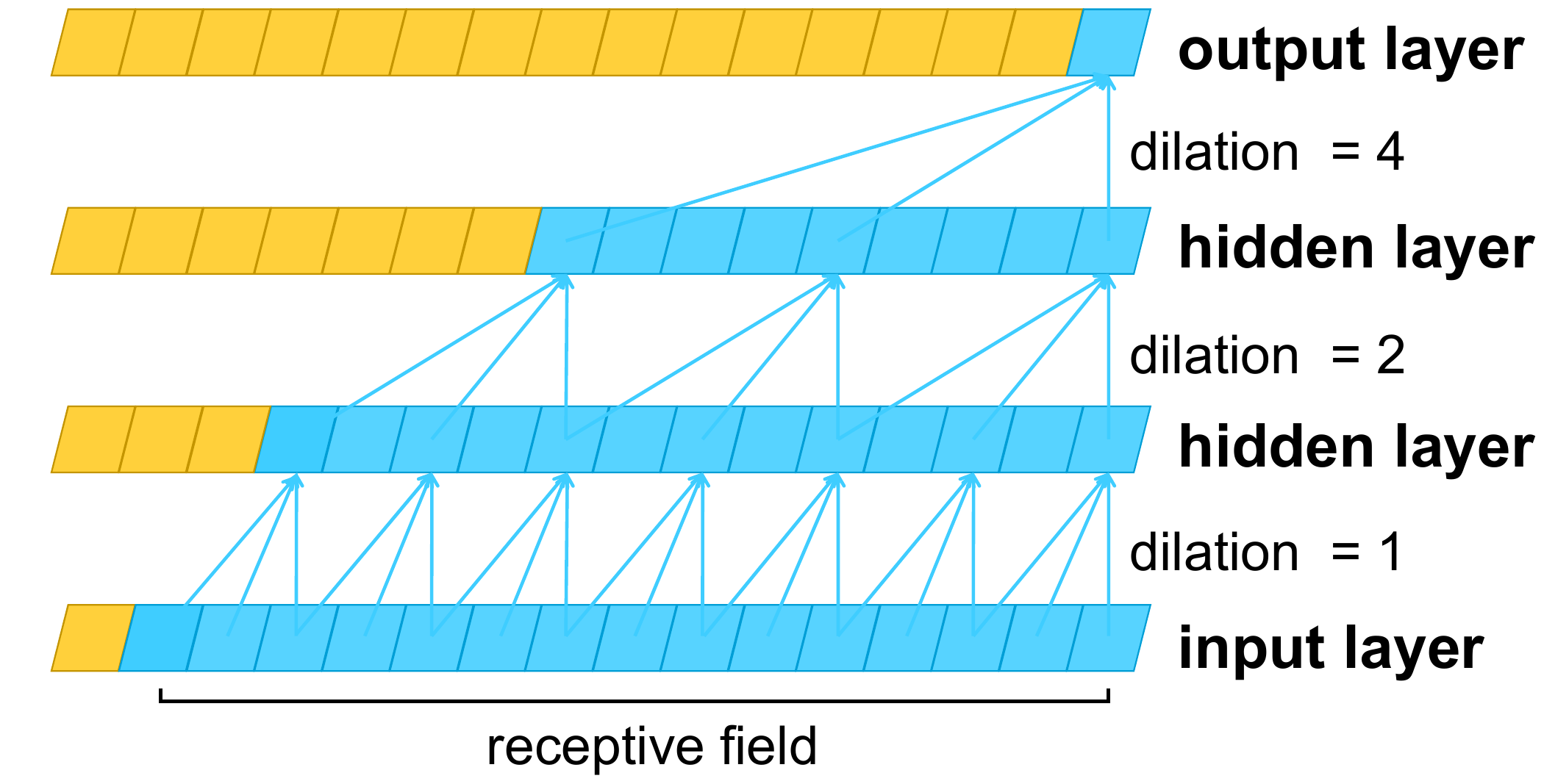}}
  \centerline{(b) Dilated causal convolutions}
 \end{minipage}
 \caption{Superiority of dilated causal convolutions compared to casual convolutions (Better viewed in color)}
 \label{fig4} 
\end{figure}

Inspired by \cite{van2016wavenet}, we use the causal convolution operation, built upon 1D convolutions, to extract the sequential information.
Specifically, it can keep the sequence order invariant and guarantee no information leakage from the future to the past by using padding and filter shifting.
A simple example is presented in Fig.~\ref{fig4}~(a), which depicts a stack of causal convolutions with kernel size $3\small{\times}1$.
However, the causal convolution usually requires very large kernel sizes or too many layers to increase the receptive field, leading to a large number of parameters.

To overcome this, inspired by \cite{bai2018empirical}, we use the dilated operation to improve the causal convolution, since it can guarantee exponentially large receptive fields~\cite{YuKoltun2016}.
To be specific, the dilation operation is equivalent to introducing a fixed step between every two adjacent filter taps~\cite{bai2018empirical}.
A simple example is provided in Fig.~\ref{fig4}~(b), which depicts a stack of dilated causal convolutions with kernel size $3\small{\times}1$.
One can find that the receptive field of the dilated causal convolution is much larger than the causal convolution.
Specifically, the gap of receptive fields between the two convolutions increases exponentially with the increase of the network depth.

\subsubsection{Correlational convolutions}
Note that existing fully convolution networks~\cite{bai2018empirical,van2016wavenet,long2015fully}, \eg, dilated causal convolutions, can hardly extract asset correlations, since they process the price of each asset separately by using 1D convolutions.
To address this, we devise a correlational convolution operation, which seeks to combine the price information from different assets, by fusing the features of all assets at every time step.
Specifically, we apply padding operations to keep the structure of feature maps invariant.
With this operation, the correlation information net can construct a multi-block architecture as shown in Fig.~\ref{fig2}~(bottom), and asymptotically extract the asset correlation without changing the structure of asset features.

In addition, we denote a degenerate variant of TCCB as TCB which does not use the correlational convolution operation.
Concretely, TCB only extracts the price sequential information by using dilated causal convolutions.
We empirically show that the correlation information net with TCCB can extract good asset correlations and helps to gain more profits, compared to TCB (See results in Section~\ref{representation62}). This result further demonstrates the significance of the asset correlation in portfolio selection and confirms the effectiveness of the correlation information net.

\subsection{Decision-making Module}

Based on all extracted feature maps, PPN makes the final portfolio decision.
To avoid heavy transaction costs, we adopt a recursive mechanism~\cite{moody2001learning} in the decision-making.
That is, the decision-making requires considering the action from last period, which helps to discourage huge changes between portfolios and thus constrains aggressive trading.

\yifan{In practice, we directly concatenate the portfolio vector from last period into feature maps. }
Here, the recursive portfolio vector $a_{t\small{-}1}\small{\in}\mathbb{R}^{m}$ also excludes the cash term, since it is risk-free and has little influence on the learning process.
We then add a fixed cash bias  into all feature maps in order to construct complete portfolios, and decide the final portfolio $a_t\small{\in}\mathbb{R}^{m+1}$ with a convolution operation via softmax function.
We highlight that the final convolution operation is analogous to making decisions by voting all feature vectors.

\begin{remark}
The recursion mechanism of PPN makes the optimal portfolio policy time-variant, i.e., it is a non-stationary portfolio selection process~\cite{gyorfi2008growth}. More critically, the well-labeled data in portfolio selection is very scarce. These challenges make PPN hard to train with supervised learning.
\end{remark}

\section{Reinforcement Learning}

Considering the complexity of portfolio selection, instead of supervised learning, we adopt reinforcement learning to optimize PPN and develop a new cost-sensitive reward function to constrain both transaction and risk costs during the optimization.

\subsection{Direct Policy Gradient Algorithm}

With the success of AlphaGo, many deep reinforcement learning algorithms have been proposed and achieve impressive performance~\cite{mnih2015human,schulman2015trust,Lillicrap2016}.
Among these reinforcement learning algorithms, the most suitable one seems to be DDPG, since it can directly approximate the deterministic portfolio decision with DNNs.
Nevertheless, DDPG needs to estimate the state-action values via a Q network, which is often hard to learn and usually fails to converge even in a simple MDP~\cite{moody2001learning}.
In our case, this issue is more serious since the decision process is non-stationary.
\yifan{Hence, the selection of reinforcement learning algorithms is non-trivial.}

\yifan{Fortunately, the sequential decision-making is an immediate reward process.
That is,} the reward of portfolio selection is immediately available.
We can directly optimize the reward function, and use the policy gradient from rewards to train PPN.
We highlight that this simple policy gradient method can guarantee at least a sub-optimal solution as follows.

\begin{prop}\label{prop1}
Let $\theta$ be the parameters of the policy network, e.g., PPN and $R$ be the reward. If the policy network is updated approximately proportional to the gradient $\triangle\theta \small{\thickapprox} \eta \frac{\partial R}{\partial \theta}$, where $\eta$ is the learning rate, then $\theta$ can usually be assured to converge to a local optimal policy in the reward $R$~\cite{sutton2000policy}.
\end{prop}

We will further discuss the selection of reinforcement learning in Section~\ref{selection7}.

\subsection{Cost-sensitive Reward Function}

To constrain both transaction and risk costs, we develop a new cost-sensitive reward function. To this end, we first devise a risk-sensitive reward regarding no transaction cost.

\subsubsection{Risk-sensitive reward}
Assuming there is no transaction cost, most existing methods use the log-return ($\log r_t$) as the reward, since it helps to guarantee a log-optimal strategy.

\begin{prop} \label{prop2}
If there is no transaction cost and the market is stationary ergodic, the portfolio policy that maximizes the \textbf{expected log-return} $\mathbb{E}\{\log r_t\}$ can achieve a \textbf{log-optimal strategy}, with the theoretical maximal growth rate $\bar{W}^*\small{=}\lim_{t\small{\rightarrow} \infty}\frac{1}{t}\log \bar{S}^*_t$, where $\bar{S}_t^*$ is the accumulated wealth~\cite{gyorfi2008growth}.
\end{prop}

In practice, we can use the empirical approximation of the expected log-return $\mathbb{E}\{\log r_t\}$ as the reward: $R =\frac{1}{T}\sum_{t=1}^T \hat{r}_t$,
where $\hat{r}_t \small{:=} \log r_t$ is the log-return on the $t$-th period, and $T$ is the total number of sampled portfolio data.

However, this reward ignores the risk cost, thus being less practical.
To solve this issue, we define the empirical variance of log-return on sampled portfolio data as the risk penalty, \ie, $\sigma^2(\hat{r}_t|t\small{=}1,..,T)$, shortly $\sigma^2(\hat{r}_t)$, and then develop a risk-sensitive reward function as:
\begin{align}
 R = \frac{1}{T}\sum_{t=1}^T \hat{r}_t - \lambda \sigma^2(\hat{r}_t), \nonumber
\end{align}
where $\lambda\geq 0$ is a trade-off hyperparameter.

We next show the near-optimality of the risk-sensitive reward. It represents the relationship between the policy regarding this risk-sensitive reward and the log-optimal strategy in Prop.~\ref{prop2} which cannot constrain the risk cost.

\begin{theorem}\label{thm1}
Let $\bar{W}^*$ be the growth rate of the log-optimal strategy and $S^*_t$ be the wealth achieved by the optimal portfolio policy that maximizes $\mathbb{E}\{\log r_t\} \small{-} \lambda Var\{\log r_t\}$. Under the same condition as in Prop.~\ref{prop2}, for any $\lambda\small{\geq} 0$ and $\frac{1}{e}\small{\leq} r_t \small{\leq} e$, the maximal growth rate of this policy satisfies:
\begin{align}
 \bar{W}^* \geq \liminf_{t\small{\rightarrow}\infty}\frac{1}{t}\log S^*_t \geq  \bar{W}^*-\frac{9}{4}\lambda. \nonumber
\end{align}
\end{theorem}

See the supplementary for the proof. From Theorem \ref{thm1}, when $\lambda$ is sufficiently small, the growth rate of the optimal strategy regarding this reward can approach the theoretical best one, \ie, the log-optimal strategy.

\subsubsection{Cost-sensitive reward}
Despite having theoretical guarantees, the risk-sensitive reward assumes no transaction cost, thus being insufficient. To solve this issue, we improve it by considering the proportional transaction cost.
In this setting, the log-return will be adjusted as $\hat{r}_t^c \small{:=} \log r^c_t \small{=}  \log r_t\small{*}(1\small{-}c_t)$. Specifically, the expected rebalanced log-return can also guarantee the optimality when facing the transaction cost.

\begin{prop} \label{prop3}
If the market is stationary and the return process is a homogeneous first-order Markov process, the policy that maximizes the \textbf{expected rebalanced log-return} $\mathbb{E}\{\log r^c_t\}$ can be \textbf{optimal} when facing transaction costs, with the maximal growth rate $\tilde{W}^*\small{=}\lim_{t\small{\rightarrow} \infty}\frac{1}{t}\log \tilde{S}^*_t$, where $\tilde{S}_t^*$ is the wealth achieved by this optimal strategy~\cite{gyorfi2008growth}.
\end{prop}

However, optimizing this rebalanced log-return cannot control transaction costs well.
To solve this, we further constrain the transaction cost proportion $c_t$.
Let $\omega_{t}\small{:=}1\small{-}c_t$ be the proportion of net wealth, and let $\psi_p$ and $\psi_s$ be  transaction cost rates for purchases and sales.
On the $t$-th period, after making the decision $a_t$, we need to rebalance from the current portfolio $\hat{a}_{t-1}\small{=}\frac{a_{t-1}\odot x_{t-1}}{a_{t-1} ^\top x_{t-1}}$ to $a_t$, where $\odot$ is the element-wise product.
During rebalancing, the sales occur if $\hat{a}_{t-1,i}\small{-}a_{t,i}\omega_{t}\small{>}0$, while the purchases occur if $a_{t,i}\omega_{t}\small{-}\hat{a}_{t-1,i}\small{>}0$. Hence,
\begin{align}
 c_t = \psi_s \sum_{i=1}^m(\hat{a}_{t-1,i}\small{-}a_{t,i}\omega_{t})^{+} \small{+} \psi_p \sum_{i=1}^m (a_{t,i}\omega_{t}\small{-}\hat{a}_{t-1,i})^{+}, \nonumber
\end{align}
where $(x)^{+}\small{=} \max(x,0)$. Following~\cite{li2017transaction}, we set $\psi_p \small{=} \psi_s\small{=} \psi\in[0,1]$, and then obtain:
\begin{align}
 c_t = \psi \|a_t\omega_{t} -\hat{a}_{t-1}\|_1. \nonumber
\end{align}
Getting rid of $\omega_t$, we can bound $c_t$ as follows.

\begin{prop}\label{prop4}
Let $\psi$ be the transaction cost rate, $\hat{a}_{t-1}$ and $a_t$ be the asset allocations before and after rebalancing. The cost proportion $c_t$ on the $t$-th period is bounded: 
\begin{align}
 \frac{\psi}{1+\psi}\|a_t\small{-}\hat{a}_{t-1}\|_1 \leq c_t \leq \frac{\psi}{1-\psi}\|a_t\small{-}\hat{a}_{t-1}\|_1, \nonumber
\end{align}
where $\|a_t\small{-}\hat{a}_{t-1}\|_1 \in \Big{(}0,\frac{2(1-\psi)}{1+\psi}\Big{]}$.
\end{prop}


See the supplementary for the proof. Prop. \ref{prop4} shows that both upper/lower bounds of $c_t$ are related to $\|a_t\small{-}\hat{a}_{t-1}\|_1$: the smaller the L1 norm, the smaller the upper/lower bounds and thus the $c_t$. By constraining this L1 norm, we derive the final cost-sensitive reward based on Theorem~\ref{thm1} and Prop. \ref{prop4} as:
\begin{align}\label{eq1}
R = \underbrace{\frac{1}{T}\sum_{t=1}^T \hat{r}_t^c - \lambda \sigma^2(\hat{r}_t^c)}_{\text{risk-sensitive reward}}-\underbrace{\frac{\gamma}{T\small{-}1}\sum_{t=2}^T \|a_t\small{-}\hat{a}_{t-1}\|_1}_{\text{transaction cost constraint}},
\end{align}
where $\gamma\geq 0$ is a trade-off hyperparameter.

We next show the near-optimality of the cost-sensitive reward, which reflects the relationship between the strategy regarding this cost-sensitive reward and the theoretical optimal strategy in Prop.~\ref{prop3} which cannot control both costs.

\begin{theorem}\label{thm3}
Let $\tilde{W}^*$ be the growth rate of the theoretical optimal strategy that optimizes $\mathbb{E}\{\log r^c_t\}$, and $S^*_t$ be the wealth achieved by the optimal policy that maximizes $\mathbb{E}\{\log r_t^c\} \small{-} \lambda Var\{\log r_t^c\} \small{-} \gamma \mathbb{E}\{\|a_t\small{-}\hat{a}_{t-1}\|_1\}$. Under the same condition as in Props.~\ref{prop3} and \ref{prop4}, for any $\lambda\small{\geq} 0$, $\gamma\small{\geq}0$, $\psi \small{\in}[0,1]$ and $\frac{1}{e}\small{\leq} r_t^c \small{\leq} e$, the maximal growth rate of this policy satisfies:
\begin{align}
 \tilde{W}^* \geq \liminf_{t\small{\rightarrow}\infty}\frac{1}{t}\log S^*_t >\tilde{W}^*-\frac{9}{4}\lambda - \frac{2\gamma(1-\psi)}{1+\psi}. \nonumber
\end{align}
\end{theorem}


See the supplementary for the proof. Specifically, when $\lambda$ and $\gamma$ are sufficiently small, the wealth growth rate of the strategy regarding the cost-sensitive reward can be close to the theoretical optimum.

We highlight that this reward can be helpful to design more effective portfolio selection methods with the near-optimality guarantee when facing both transaction and risk costs in practice tasks.
Specifically, by optimizing this reward with the direct policy gradient method, the proposed PPN can learn at least a sub-optimal policy to effectively maximize accumulated returns while controlling both costs as shown in Prop.~\ref{prop1}.

\begin{remark}
The denominator $T$ in Eqn. (\ref{eq1}) ensures that the rewards from different price sequences are equivalent. Moreover, the assumption (ii) makes the action and environment isolated, allowing us to use the same price segment to evaluate different actions. These enable us to train PPN with the online stochastic batch method~\cite{jiang2017deep}, which helps to improve the data efficiency.
\end{remark}

\section{Experimental Results}

We evaluate PPN in three main aspects: (1) the profitability on real-world datasets; (2) the feature extraction ability for portfolio series; (3) the cost-sensitivity to both transaction and risk costs. To this end, we first describe the baselines, metrics, datasets and implementation details in experiments.

\subsection{Experimental Settings} 
\subsubsection{Baselines}
We compare PPN with several state-of-the-art methods, including Uniform Buy-And-Hold (UBAH), best strategy in hindsight (Best), CRP~\cite{cover1991universal}, UP~\cite{cover1991universal}, EG~\cite{helmbold1998line}, Anticor~\cite{borodin2004can}, ONS~\cite{agarwal2006algorithms},  CWMR~\cite{li2011confidence}, PAMR~\cite{li2012pamr}, OLMAR~\cite{li2012line}, RMR~\cite{huang2013robust}, WMAMR~\cite{gao2013weighted} and EIIE~\cite{jiang2017deep}. In addition, to evaluate the effectiveness of the asset correlation, we also compare PPN with a degenerate variant PPN-I that only exploits independent price information by using TCB.

\subsubsection{Metrics}
Following~\cite{shen2017portfolio,li2017transaction}, we use three main metrics to evaluate the performance.
The first is  \emph{accumulated portfolio value} (APV), which evaluates the profitability when considering the transaction cost.
\begin{equation}
  \text{APV} = S_n = S_0 \prod_{t=1}^n a_t^{\top}x_t(1\small{-}c_t), \nonumber
\end{equation}
where $S_0=1$ is the initialized wealth.
In addition, $a_t$, $x_t$ and $c_t$ indicate the portfolio vector, the  price relative vector and  the transaction cost proportion on the $t$-th round, respectively.

A major drawback of APV is that it neglects the risk factor.
That is, it only relies on the returns without considering the fluctuation of these returns.
Thus, the second metric is  \emph{Sharpe Ratio} (SR), which evaluates the average return divided by the fluctuation, \ie, the \emph{standard deviation} (STD) of returns.
\begin{equation}
  \text{SR} = \frac{\text{Average}(r_t^c)}{\text{Standard Deviation}(r_t^c)}, \nonumber
\end{equation}
where $r_t^c$ is the rebalanced log-return on the $t$-th round.

Although SR considers the volatility of portfolio values, it treats upward and downward movements equally. However, downward movements are usually more important, since it measures algorithmic stability in the market downturn.
To highlight the downward deviation, we further use \emph{Calmar Ratio}~(CR), which measures the accumulated profit divided by \emph{Maximum Drawdown} (MDD):
\begin{equation}
  \text{CR} = \frac{S_n}{\text{MDD}}, \nonumber
\end{equation}
where MDD denotes the biggest loss from a peak to a trough:
\begin{equation}
\text{MDD} = \max_{t: \tau > t}  \frac{S_t - S_{\tau}}{S_t}. \nonumber
\end{equation}
  
Note that the higher APV, SR and CR values, the better profitability of algorithms; while the lower STD and MDD values, the higher stability of returns.
We also evaluate average \emph{turnover} (TO) when examining the influence of transaction costs, since it estimates the average trading volume.
\begin{equation}
  \text{TO} = \frac{1}{2n}\sum_{t=1}^n\|\hat{a}_{t-1}\small{-}a_{t}\omega_{t}\|_1, \nonumber
\end{equation}
where $\hat{a}_{t-1}$ and $\omega_{t}$ indicate the current portfolio before rebalance and net wealth proportion   on the $t$-th round.

\subsubsection{Datasets and preprocessing} 
The globalization and the rapid growth of crypto-currency markets yield a large number of data in the finance industry.
Hence, we evaluate PPN on several real-world crypto-currency datasets.
Following the data selection method in~\cite{jiang2017deep}, all datasets are accessed with Poloniex\footnote{Poloniex's official
API: https://poloniex.com/support/api/.}. 
To be specific, we set the bitcoin as the risk-free cash and select risk assets according to the crypto-currencies with top month trading volumes in Poloniex.
We summary statistics of the datasets in Table~\ref{table1}.
All assets, except the cash asset, contain all 4 prices.
The price window of each asset spans 30 trading periods, where each period is with 30-minute length.

Some crypto-currencies might appear very recently, containing some missing values in the early stage of data.
To fill them, we use the flat fake price-movements method~\cite{jiang2017deep}.
Moreover, the decision-making of portfolio selection relies on the relative price change rather than the absolute change~\cite{li2012line}.
We thus normalize the price series with the price of the last period.
That is, the input price on the $t$-th period is normalized by $P_t \small{=} \frac{P_t}{P_{t,30}} \small{\in}\mathbb{R}^{m\times 30 \times 4}$, where $P_{t,30} \small{\in} \mathbb{R}^{m \times 4}$ represents the prices of the last period.

\begin{table}[h] 
\caption{The detailed statistics information of the used crypto-currency datasets} 
 \label{table1}
 \begin{center}
 \begin{scriptsize}
 \scalebox{0.92}{
 \renewcommand*\arraystretch{1}
  \begin{tabular}{|c|c|c|c|c|c|}\hline
  \multirow{2}{*}{\hspace{-0.01in}Datasets\hspace{-0.01in}} & \multirow{2}{*}{\hspace{-0.047in} $\#$Asset \hspace{-0.047in}}&\multicolumn{2}{c|}{Training Data}&\multicolumn{2}{c|}{Testing Data}\cr
  \cline{3-6}
  & & Data Range& \hspace{-0.065in} Num. \hspace{-0.065in} & Data Range& \hspace{-0.06in} Num. \hspace{-0.06in} \cr
  \hline
       Crypto-A   &  12 & 2016-01 to 2017-11   & 32269  &   2017-11 to 2018-01 	&  2796    \\
       Crypto-B  & 16  & 2015-06 to 2017-04   &   32249 	&   2017-04 to 2017-06  	 	&  2776 	\\
       Crypto-C   & 21  & 2016-06 to 2018-04  &  32205   &   2018-04 to 2018-06	    &  2772  	 \\
       Crypto-D   &  44 & 2016-08 to 2018-06 & 32205  &  2018-06 to 2018-08   & 2772\\
  \hline%
  \end{tabular}}
 \end{scriptsize}
 \end{center}
\end{table} 


\begin{table*}[t]
\vspace{0.1in}
	\caption{Detailed network architecture of the proposed portfolio policy network, where we use the following abbreviations: CONV: convolution layer; N: the number of output channels; K: kernel size; S: stride size; P: padding size or operation name; DiR: dilation rate; DrR: dropout rate }
	\label{table:architecture}
	\centering
		\begin{tabular}{c|c|c}
			\hline
			\hline
			\multicolumn{3}{c}{\textbf{Correlation Information Net}} \\
			\hline
			\hline
			{Part} & {Input $ \rightarrow $ Output shape} & {Layer information} \\
			\hline
			\multirow{3}[0]{*}{TCCB1}
			& $ (m, k, 4) \rightarrow (m, k, 8) $ & DCONV-(N8, K[1x3], S1, P2), DiR1, DrR0.2, ReLU \\	
			& $ (m, k, 8) \rightarrow (m, k, 8) $ & DCONV-(N8, K[1x3], S1, P2), DiR1, DrR0.2, ReLU \\
			& $ (m, k, 8) \rightarrow (m, k, 8) $ & CCONV-(N8, K[mx1], S1, P-SAME), DrR0.2,  ReLU  \\
			\hline
			\multirow{3}[0]{*}{TCCB2}
			& $ (m, k, 8) \rightarrow (m, k, 16) $ & DCONV-(N16, K[1x3], S1, P4), DiR2, DrR0.2, ReLU\\
			& $ (m, k, 16) \rightarrow (m, k, 16) $ & DCONV-(N16, K[1x3], S1, P4), DiR2, DrR0.2, ReLU \\
			& $ (m, k, 16) \rightarrow (m, k, 16) $ & CCONV-(N16, K[mx1], S1, P-SAME), DrR0.2,  ReLU  \\
			\hline
			\multirow{3}[0]{*}{TCCB3}
			& $ (m, k, 16) \rightarrow (m, k, 16) $ & DCONV-(N16, K[1x3], S1, P8), DiR4, DrR0.2, ReLU\\	
			& $ (m, k, 16) \rightarrow (m, k, 16) $ & DCONV-(N16, K[1x3], S1, P8), DiR4, DrR0.2, ReLU\\
			& $ (m, k, 16) \rightarrow (m, k, 16) $ & CCONV-(N16, K[mx1], S1, P-SAME), DrR0.2,  ReLU  \\
			\hline
			\multirow{1}[0]{*}{Conv4}
			& $ (m, k, 16) \rightarrow (m, 1, 16) $ & CONV-(N16, K[1xk], S1, P-VALID),   ReLU  \\
			\hline
			\hline
			\multicolumn{3}{c}{\textbf{Sequential Information Net}} \\
			\hline
			\hline
			\multirow{1}[0]{*}{LSTM}
			& $ (m, k, 4) \rightarrow (m, 1, 16) $ & LSTM unit number:16 \\
            \hline
            \hline
			\multicolumn{3}{c}{\textbf{Decision-making Module}} \\
            \hline
            \hline
			\multirow{1}[0]{*}{Concatenation}
			& $ (m, 16)\small{\oplus}(m,  16)\small{\oplus}(m, 1) \small{\oplus}(1,33) \rightarrow (m+1, 33) $ & Concatenation of extracted features and other information \\
			\hline
			\multirow{1}[0]{*}{Prediction}
			& $ (m+1, 33) \rightarrow (m+1, 1) $ & CONV-(N1, K[1x1], S1, P-VALID),   Softmax \\
			\hline
			\hline
		\end{tabular}
\end{table*}

\subsubsection{Implementation details}\label{implementation}

As mentioned above, the overall network architecture of PPN is shown in Fig.~\ref{fig2}.
Specifically, there are three main components: Correlation Information Net, Sequential Information Net and Decision-making Module.
To make it more clearer, we record the detailed architectures in Table~\ref{table:architecture}.

To be specific, in Correlation Information Net, we adopt the temporal correlational convolutional block as the basic module.
To be specific, it consists of two components, \ie, the dilated causal convolution layer (DCONV) and the correlation convolution layer~(CCONV).

Note that the concatenation operation in the decision-making module has two steps. First, we concatenate all extracted features and the portfolio vector from last period. Then, we concatenate the cash bias into all feature maps.

In addition, we implement the proposed portfolio policy network with Tensorflow~\cite{tensorflow}.  Specifically, we use Adam optimizer with batch size 128 on a single NVIDIA TITAN X GPU. We set the learning rate to 0.001, and choose $\gamma$ and $\lambda$ from $10^{[-4:1:-1]}$ using cross-validations. Besides, the training step is $10^5$, the cash bias is fixed to 0, and the transaction cost rate is $0.25\%$, which is the maximum rate at Poloniex.
In addition, the training time of PPN is about 4, 5.5, 7.5 and 15 GPU hours on the Crypto-A, Crypto-B, Crypto-C and Crypto-D datasets, respectively. 
All results on crypto-currency datasets are averaged over 5 runs with random initialization seeds.

\begin{table*}[t]
\caption{Performance comparisons on different datasets}
 \label{table2}
 \begin{center}
 \scalebox{1}{
	\begin{tabular}{|l|c|c|c|c|c|c|c|c|c|c|c|c|c|c|c|}\hline
  \multirow{2}{*}{Algos} & \multicolumn{3}{c|}{Crypto-A}&&\multicolumn{3}{c|}{Crypto-B}&&\multicolumn{3}{c|}{Crypto-C}&&\multicolumn{3}{c|}{Crypto-D}\cr
  \cline{2-16}
  & APV & SR($\small{\%}$) & CR && APV & SR($\small{\%}$) &CR && APV & SR($\small{\%}$) &CR && APV & SR($\small{\%}$) &CR \cr
  \hline
      UBAH & 2.59 & 3.87 & 3.39          &&1.63 & 2.57 & 2.43              && 1.32 & 3.00 & 1.61   && 0.63 & 0.20 & -5.85\\\hline
      Best &6.65 & 4.59 & 13.67         &  & 3.20 & 2.95 & 3.61             &&2.97 & 3.15 & 3.96   && 1.04 & 0.64 & 0.63\\\hline
      CRP &2.40 & 3.95 & 3.24           &&1.90 & 3.77 & 3.81            && 1.30 & 3.30 & 1.77      && 0.66 & 0.20 & -5.85\\\hline
      UP & 2.43  & 3.95 & 3.28           &  &1.89 & 3.70 &3.71        &&1.30 & 3.29 & 1.76         && 0.66 & 0.20 &  -5.85\\\hline
      EG & 2.42 & 3.96 & 3.27            & &1.89 & 3.71 & 3.75            && 1.30 & 3.30 & 1.76    && 0.67 & 0.21 &  -5.85\\\hline
      Anticor &2.17 & 2.96 & 2.23         & &21.80 & 9.92 & 103.68          && 0.75 & -1.48 & -0.91    &&  3.14 & 6.81 & 66.28\\\hline
      ONS & 1.28  & 1.40 &  0.53           &&1.71 & 3.15 & 4.00             && 1.14 & 2.33 & 1.95      && 1.00 & 0.19 & 0.01\\\hline
      CWMR & 0.01 & -8.21 & -0.99        &&0.64 & 0.42 & -0.54            && 0.01 & -16.61 & -0.99     && 0.38 & -0.02 &  -5.19\\\hline
      PAMR & 0.01 & -7.17 & -0.99        &&0.880 & 0.91 & -0.20              && 0.01 & -15.48 & -0.99   && 0.82 & 0.08 & -1.78\\\hline
      OLMAR &0.65 & 0.32 & -0.47          &&774.47 & 10.91 & 2040.70             && 0.05 & -7.56 & -0.99       && 11.25 & 7.21 & 135.97\\\hline
      RMR & 0.69 & 0.46  & -0.42             && 842.26 & 11.62 & 3387.69         &&0.05 & -7.72 & -0.99        && 14.337& 7.93 & 192.59\\\hline
      WMAMR & 0.85 & 0.67 &  -0.22          &&87.85 & 8.25 & 245.23                 && 0.26 & -3.78 & -0.98    && 7.72 & 6.62 &  227.16\\\hline
      EIIE &10.48  & 5.27 & 21.47          && 2866.15 & 13.42 & 8325.78           && 2.87 & 4.04 & 9.54        && 113.58 & 15.11 & 4670.91\\\hline
      PPN-I~(ours)& 25.76  & 6.75 & 57.05        && 7549.35 &  14.74 & 28915.43    && 3.93 & 5.12 & 15.04      &&  238.93 & 16.07 & 8803.95\\\hline
      PPN~(ours) & \textbf{32.04} & \textbf{6.85} &\textbf{79.87}          &&\textbf{9842.56} & \textbf{14.82}  &\textbf{37700.03}            &&\textbf{4.81} & \textbf{5.89} & \textbf{16.11}  && \textbf{538.22} & \textbf{17.82} & \textbf{15875.72}\\\hline
  \end{tabular}}
 \end{center}
\end{table*}
\subsection{Evaluation on Profitability}
We first evaluate the profitability of PPN, and record the detailed performance in Table~\ref{table2}.

From the results, EIIE and PPN-based methods perform better than all other baselines in terms of APV.
Since the three methods adopt neural networks to learn policies via reinforcement learning, this observation demonstrates the effectiveness and superiority of deep reinforcement learning in portfolio selection.

Moreover, PPN-based methods perform better than EIIE.
This finding implies that PPN-based methods can extract better sequential feature representation, which helps to learn more effective portfolio policies with better profitability.

In addition, PPN outperforms PPN-I in terms of APV.
This observation confirms the effectiveness and significance of the asset correlation in portfolio selection.

Last, PPN also achieves the best or relatively good SR and CR performance.
Since both metrics belong to risk-adjusted metrics, this finding implies that PPN is able to gain more stable profits than other baselines.

\begin{table*}[t]
	\caption{Evaluations of portfolio policy network with different feature extractors}
	\vspace{-0.1in}
 \label{table3}
 \begin{center}
 \scalebox{0.87}{
 \begin{small}
	\begin{tabular}{|c|c|c|c|c|c|c|c|c|c|c|c|c|c|c|c|}\hline
  \multirow{2}{*}{Module} & \multicolumn{3}{c|}{Crypto-A}&&\multicolumn{3}{c|}{Crypto-B}&&\multicolumn{3}{c|}{Crypto-C}&&\multicolumn{3}{c|}{Crypto-D}\cr \cline{2-16}
	   & APV & SR($\small{\%}$) & CR && APV & SR($\small{\%}$) & CR && APV & SR($\small{\%}$) & CR && APV & SR($\small{\%}$) & CR   \\\hline
      PPN-LSTM &  14.48  &  5.62 & 38.19   &&      3550.32 & 13.75 &13297.32               && 2.85 & 3.99 & 6.69                   && 159.54  &  15.16 & 6319.84                        \\\hline
      PPN-TCB &12.76 &  5.40  &26.52       &&      3178.42 & 13.63 &  11011.87                 &&         2.01 & 3.32 & 4.66             && 102.85  &  14.09 & 2972.63                 \\\hline
      PPN-TCCB & 16.51 &  6.01 & 35.89      &&     4181.17 & 13.85 & 15798.05                 &&       3.29 & 4.53 & 12.38                && 171.82  &  14.97 & 3945.99                     \\\hline
      PPN-TCB-LSTM & 18.62 & 6.28 & 39.87   &&      4485.89 &14.18 & 15232.31                  &&      3.49 & 4.48 &10.96                    && 179.43  & 15.18& 9150.33                      \\\hline
      PPN-TCCB-LSTM &  21.03  &   6.12 & 52.75  &&   5575.25 &  14.46 & 21353.23                  &&     3.69 & 4.72 &   10.50                  && 224.41  &  15.99 &  8522.43                      \\\hline
      PPN-I & 25.76&  6.75   &57.05     &&       7549.35 & 14.74 &  28915.43              &&          3.93 & 5.12 & 15.04           && 238.93  & 16.07 & 8803.95           \\\hline
      PPN &   \textbf{32.04} & \textbf{6.85} &  \textbf{79.87}   && \textbf{9842.56} & \textbf{14.82}  &\textbf{37700.03}      &&\textbf{4.81} & \textbf{5.89} & \textbf{16.11}         && \textbf{538.22} & \textbf{17.82} & \textbf{15875.72}    \\\hline
	\end{tabular}
 \end{small}}
 \end{center}
 \vspace{-0.1in}
\end{table*}

\subsection{Evaluation on Representation Ability}\label{representation62}

\begin{figure}[t]
  \includegraphics[width=8.8cm]{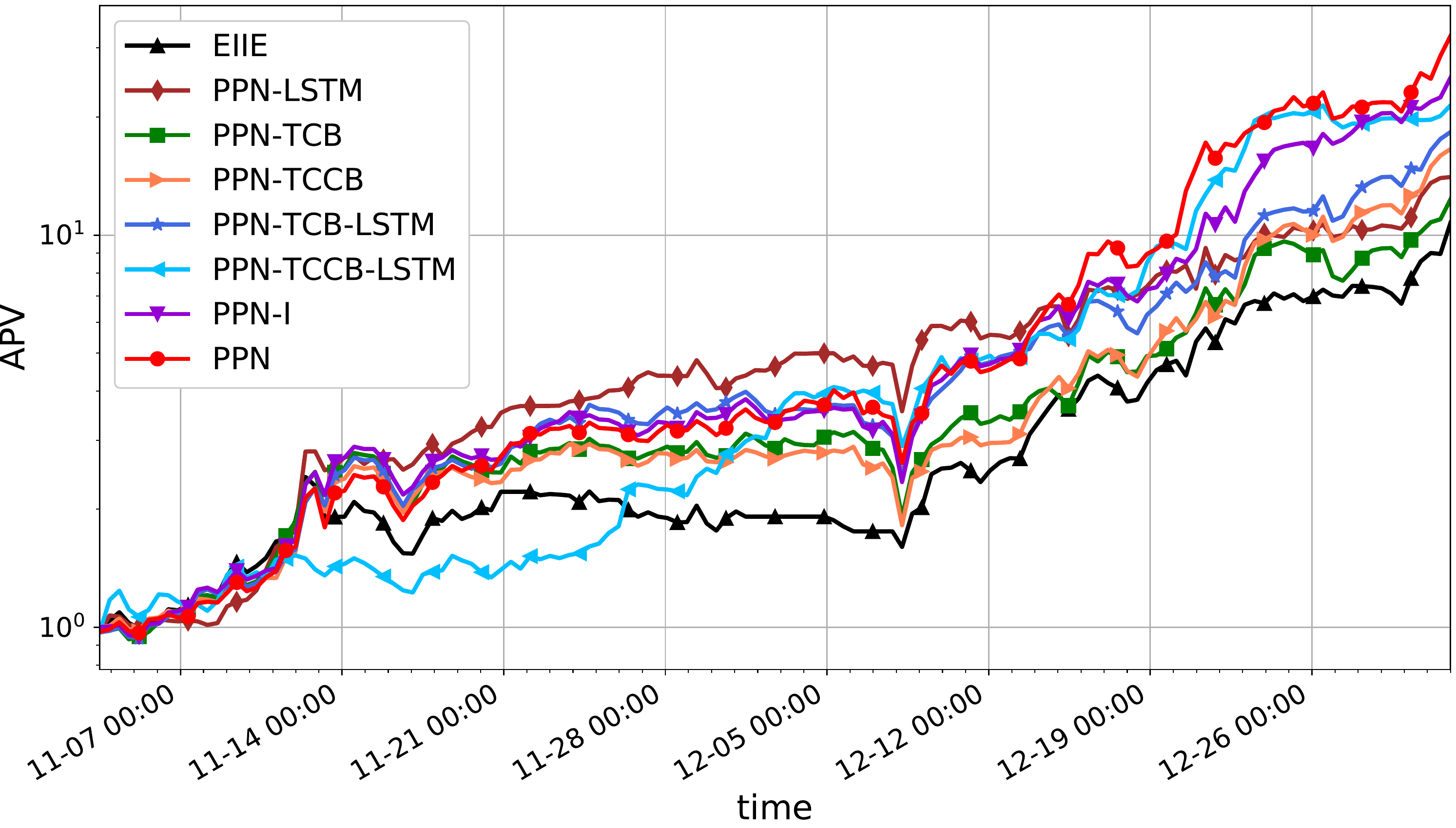}
  \caption{\blue{The performance development of the proposed portfolio policy network with different feature extractors and EIIE on the Crypto-A dataset (Better viewed in color). A larger scale version of this figure can be found in Appx. D.1}}
  \label{fig5}
  \vspace{-0.1in}
\end{figure}

We next evaluate the representation abilities of PPN with different extraction modules, when fixing all other parameters.
Specifically, we compare PPN and PPN-I with the variants that only adopt one module, \ie, LSTM, TCB or TCCB, namely PPN-LSTM, PPN-TCB and PPN-TCCB.
To demonstrate the parallel structure, we also compare PPN and PPN-I with the variants that use the cascaded structure, namely PPN-TCB-LSTM and PPN-TCCB-LSTM.
The only difference among these variants is that the extracted features are different.
We  present the results in Table~\ref{table3} and Fig.~\ref{fig5}, from which we draw several observations.

Firstly, we discuss the variants that only use one feature extraction module.
Specifically, PPN-LSTM outperforms PPN-TCB, which means the proposed sequential information net extracts better price sequential patterns.
Besides, PPN-TCCB outperforms PPN-LSTM and PPN-TCB, which verifies both TCCB and the correlation information net.

Secondly, all \yifan{variants} that consider asset correlations, \ie, PPN, \yifan{PPN-TCCB and PPN-TCCB-LSTM, outperform their independent variants, \ie, PPN-I, PPN-TCB and PPN-TCB-LSTM}. This observation confirms the  significance and effectiveness of the asset correlation in portfolio selection.

Thirdly, all combined variants, \ie, PPN, PPN-I and cascaded modules, outperform the variants that only adopt LSTM, TCB or TCCB. This means that combining both types of information helps to extract better features, which further confirms the effectiveness of the two-stream architecture.

Next, PPN outperforms all other \yifan{variants}, which confirms its strong representation ability.
Note that PPN is not always the best throughout the backtest in Fig.~\ref{fig5}.
For example, in the early stage, many \yifan{variants} perform similarly. But in the late stage, PPN performs very well.
Considering that the correlation between two price events decreases exponentially with their sequential distance~\cite{holt2004forecasting}, this result demonstrates better generalization abilities of PPN.

\blue{Lastly, as shown in Fig.~\ref{fig5}, there are some periods that all methods (EIIE and PPN based methods) suffer from significant draw-down, like in the middle November and the earlier December. Since it is model-agnostic, such draw-down may result from the market factor instead of the methods themselves. Motivated by this, it is interesting to explore the market influence based on social text information  for better portfolio selection in the future.}

\begin{table*}[t]
\vspace{0.1in}
	\caption{Comparisons under different transaction cost rates on the Crypto-A dataset}
 \label{table4}
 
 \begin{center}
 \scalebox{0.95}{
 \begin{small}
	\begin{tabular}{|c|c|c|c|c|c|c|c|c|c|c|c|c|c|c|c|c|}\hline

   \multirow{2}{*}{Algos}&  \multicolumn{2}{c|}{c=0.01$\%$}& \multicolumn{2}{c|}{c=0.05$\%$} & \multicolumn{2}{c|}{c=0.1$\%$} & \multicolumn{2}{c|}{c=0.25$\%$}& \multicolumn{2}{c|}{c=1$\%$}& \multicolumn{2}{c|}{c=2$\%$}& \multicolumn{2}{c|}{ c=5$\%$}\cr
   \cline{2-15}
   & APV & TO & APV & TO& APV & TO& APV & TO& APV & TO& APV & TO& APV & TO \cr \hline
       EIIE & 871.18  & 1.232  & 254.73   & 1.076 & 77.79 & 0.859    & 10.48 & 0.471   & 1.07 & 0.247  & 0.81  & 0.021  & \textbf{0.28} & 0.020  \\\hline
       PPN-I& 1571.67 & 0.964 & 570.73 & 0.779 & 219.30 & 0.668 &25.76 & 0.316 & 1.18 & 0.040 &0.96 & 0.013 & 0.99 & 2e-7 \\\hline
       PPN & \textbf{3741.13}  & 1.018 & \textbf{754.57} & 0.731  & \textbf{242.27} & 0.658  &\textbf{32.04} & 0.368  & \textbf{1.61} & 0.063  & \textbf{1.09} & 0.019  & \textbf{1.00} & 5e-8 \\\hline
	\end{tabular}
 \end{small}}
 \end{center}
\end{table*}

\subsection{Evaluation on Cost-sensitivity}

\subsubsection{Influences of transaction costs}
In previous experiments, we have demonstrated the effectiveness of PPN, where the transaction cost rate is $0.25\%$.
However, the effect of the transaction cost rate has not been verified. We thus examine their influences on three dominant methods on Crypto-A.

From Table~\ref{table4}, PPN achieves the best APV performance across a wide range of transaction cost rates. This observation further confirms the profitability of PPN.

Compared to EIIE, PPN-based methods obtain relatively low TO, \ie, lower transaction costs.
Since EIIE optimizes only the rebalanced log-return, this finding indicates that our proposed reward controls the transaction cost better.

Moreover, when the transaction cost rate is very large, \eg, $c\small{=}5\%$, PPN-based algorithms tend to stop trading and make nearly no gains or losses, while EIIE, however, loses most of the wealth with relatively high TO. This implies our proposed methods are more sensitive to the transaction cost.

\subsubsection{Cost-sensitivity to transaction costs}
We further evaluate the influences of $\gamma$ in the cost-sensitive reward.
From Table~\ref{table5}, we find that with the increase of $\gamma$, TO values of PPN decrease.
This observation means that when introducing $\|a_t\small{-}\hat{a}_{t-1}\|_1$ into the reward, PPN can better control the trading volume, and thus better overcome the negative effect of the transaction cost.

\begin{figure}[t]
  \centerline{\includegraphics[width=7.8cm]{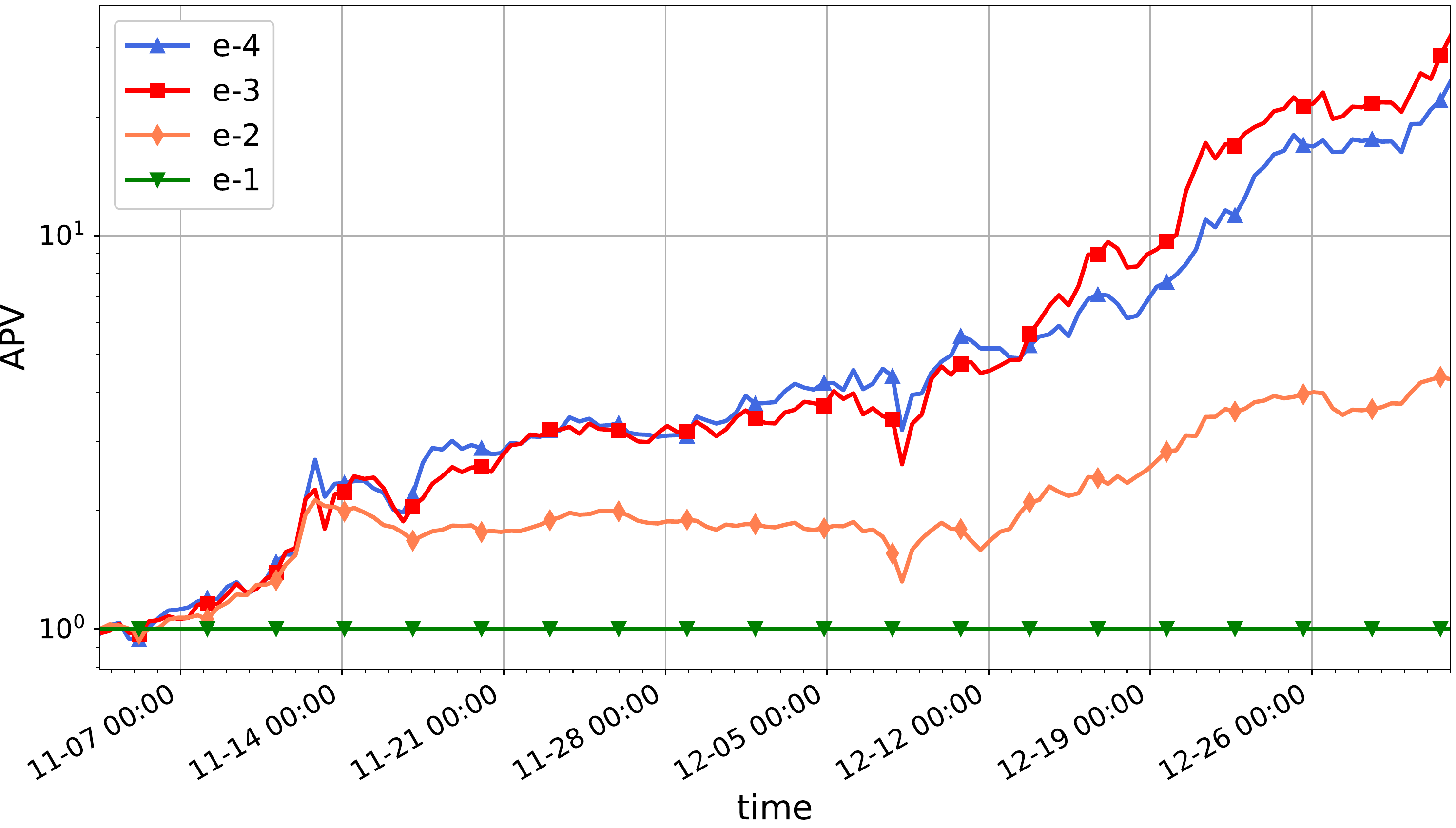}} 
  \caption{\blue{Performance development of portfolio policy network under different $\gamma$ on the Crypto-A dataset.  A larger scale version of this figure can be found in Appx. D.1}}
  \label{fig6}
 \vspace{-0.1in}
\end{figure}

This finding is also reflected in Fig.~\ref{fig6}.
With the increase of $\gamma$, there are more period intervals remaining unchanged. That is, when the transaction cost outweighs the benefit of trading, PPN will stop the meaningless trading.

Note that, PPN achieves the best APV performance when $\gamma\small{=}10^{-3}$ in Table~\ref{table5}.
This observation is easy to understand. If $\gamma$ is too small, \eg, $10^{-4}$, PPN tends to trade aggressively, leading to a large number of transaction costs, thus affecting the profitability of PPN.
If $\gamma$ is large, \eg, $10^{-2}$ and $10^{-1}$, PPN tends to trade passively, thus limiting the model to seeking better profitability.
As a result, when setting \yifan{a more reasonable value}, \eg, $\gamma=10^{-3}$, PPN can learn a better portfolio policy and achieve a better trade-off between the profitability and transaction costs.

\begin{table}[t]
	\caption{The performance of portfolio policy network under different $\gamma$} 
 \label{table5}
 \begin{center}
 \scalebox{0.81}{
    \begin{small}
	\begin{tabular}{|c|c|c|c|c|c|c|c|c|}\hline
    \multirow{2}{*}{$\gamma$} & \multicolumn{2}{c|}{Crypto-A}&\multicolumn{2}{c|}{Crypto-B}&\multicolumn{2}{c|}{Crypto-C}&\multicolumn{2}{c|}{Crypto-D}\cr \cline{2-9}
		 & APV  & TO &  APV  & TO & APV  & TO & APV  & TO  \\\hline
          10$^{-4}$ & 25.24  &0.433    &      2080.69  &0.950    &        4.65  &0.667    &        268.63  & 1.104 \\\hline
          10$^{-3}$ &\textbf{32.04}   &0.368 &   \textbf{9842.56}   &0.888 &      \textbf{4.81}   &0.304 &     \textbf{538.22}   &0.839  \\\hline
          10$^{-2}$ & 4.30 &   0.025   &      44.01   &0.161  &     1.21   &0.008 &     1.72   &0.012 \\\hline
          10$^{-1}$ & 1.01 & 2e-08     &      1.65   &3e-03 &     1.00   &1e-7 &     1.00  &3e-7 \\\hline
	\end{tabular}
    \end{small}}
 \end{center}
\end{table}

\begin{table*}[t]
	\caption{The performance of portfolio policy network under different $\lambda$}
\vspace{-0.1in}
 \label{table6}
 \begin{center}
 \scalebox{0.88}{
 \begin{small}
    \begin{tabular}{|c|c|c|c|c|c|c|c|c|c|c|c|c|c|c|c|c|}\hline
   \multirow{2}{*}{$\lambda$} &\multicolumn{3}{c|}{Crypto-A}&&\multicolumn{3}{c|}{Crypto-B}&&\multicolumn{3}{c|}{Crypto-C}&&\multicolumn{3}{c|}{Crypto-D}\cr \cline{2-16}
		& APV &STD(\small{\%})& MDD(\small{\%}) &&   APV &  STD(\small{\%})& MDD(\small{\%}) && APV &STD(\small{\%})  &MDD(\small{\%}) &&  APV & STD(\small{\%})& MDD(\small{\%}) \\\hline
          10$^{-4}$ & 32.04  & 2.16   & 38.86 &&      9842.56  & 2.43   &  26.11  &&      4.81  & 1.06   & 23.66 &&       538.22  & 1.32   &  20.30   \\\hline
          10$^{-3}$ & 25.56 & 1.99   & 37.40  &&      8211.08  & 2.39   & 26.11 &&      4.57   & 1.01   &  21.86 &&      300.12  & 1.28   & 20.39  \\\hline
          10$^{-2}$ & 25.38 & 1.95   & 37.26  &&      4800.81  &  2.31   & \textbf{26.10} &&      2.42  &  1.00   & 21.60  &&       264.79  &1.27    &18.43   \\\hline
          10$^{-1}$ & 9.81 & \textbf{1.85}    & \textbf{31.64} &&     3353.55  & \textbf{2.30}   & \textbf{26.10} &&      2.39  &\textbf{0.97}   & \textbf{20.12} &&    195.75    &\textbf{1.17}   & \textbf{16.54} \\
    \hline
	\end{tabular}
 \end{small}}
 \end{center}
\end{table*}

\begin{table*}[t]
\caption{\blue{Performance Comparisons on the S$\&$P500 dataset}}
\vspace{-0.1in}
 \label{table_new}
 \begin{center}
 \begin{small}
 \scalebox{0.8}{
	\begin{tabular}{|c|c|c|c|c|c|c|c|c|c|c|c|c|c|c|c|}\hline
		Algos& UBAH & Best& CRP & UP  & EG & Anticor & ONS & CWMR &PAMR & OLMAR &RMR & WMAMR & EIIE & PPN-I & PPN\\\hline \hline 
           APV  &1.21 & 89.63 &1.21& 1.22&1.22 & 1.09 &1.41 &0.99 & 1.34 &17.81&  17.77 &1.02 & 99.35 &129.27& \textbf{167.84} \\
           SR($\small{\%}$)   &12.03 & 10.37 &11.69& 11.70 &11.73 & 5.85 &36.34 &-7.75 & 8.09 &78.04& 79.29 &2.32 &  108.81 &115.27& \textbf{148}  \\
           CR  &2.39 & 1417.49 &2.34& 2.32 &2.34 &0.89 &3.86 &-0.98 &3.68 &208.32& 207.84&0.16 & 1994.89 &2135.87& \textbf{6792.51}   \\
           TO &0.021 & 0.022 & 0.033& 0.034 & 0.033 & 0.099 &0.291  &0.011 &0.115 &1.972& 1.970 &0.119 & 1.925 &1.918& 1.920 \\ 
          \hline  
	\end{tabular}}
 \end{small}
 \end{center}
 \vspace{-0.2in}
\end{table*}

\subsubsection{Cost-sensitivity to risk costs}
We also examine the influences of $\lambda$ in the cost-sensitive reward, and report the results in Table~\ref{table6}.
Specifically, with the increase of $\lambda$, the STD values of PPN asymptotically decrease on all datasets.
Since $\lambda$ controls the risk penalty $\sigma^2(\hat{r}_t)$, this result is consistent with the expectation and also demonstrates the effectiveness of PPN in controlling the risk cost.
Moreover, with the increase of $\lambda$, the MDD results decrease on most datasets.
Since MDD depends on the price volatility of the financial market, this result implies that constraining the volatility of returns is helpful to control the downward risk.

\section{Discussion}
\blue{In this section, we further discuss the architecture design of PPN, the selection of reinforcement learning algorithms for PPN and the generalization abilities of PPN.}

\subsection{Architecture Design of Portfolio Policy Network}

As shown in Fig.~\ref{fig2}, we propose a two-stream architecture for PPN. Concretely, the sequential information net is based on LSTM, and the correlation information net is built upon TCCB.
Noting that the fully convolution networks (e.g., TCB and TCCB) can also extract price sequential patterns, one may ask why we still use LSTM.

To be specific, although TCB and TCCB can learn sequential information, they can hardly make full use of them.
Specifically, the traditional convolution assumes time invariance, and uses time-invariant filters to combine convolutional features.
This makes fully convolution networks hard to extract large-scale sequence order information~\cite{bradbury2016quasi}.

As shown in Fig.~\ref{fig2}, the last Conv4  layer of the correlation information net directly uses the time-invariant filter to combine features, and hence only extracts some local sequential information.
This makes PPN-TCB perform inferior to PPN-LSTM in Table \ref{table3}.
We thus exploit LSTM to better extract the sequential representation.

On the other hand, we propose TCCB to effectively extract the asset correlation.
Such information is beneficial to improve the profitability of PPN and makes PPN-TCCB outperform PPN-TCB and PPN-LSTM in Table \ref{table3}.

In addition, note that combining both types of information can further strengthen the feature representation of portfolios and make more profitable decisions (See results in Section \ref{representation62}).
Hence, we devise a two-stream network architecture for PPN to better learn the portfolio series.

\subsection{Reinforcement Learning Algorithm Selection}\label{selection7}
We next discuss the selection of reinforcement learning algorithms. Since we use the direct policy gradient (DPG) method, we mainly discuss why not use Actor-Critic (AC) policy gradient methods.

AC requires learning a ``critic" network to approximate the value function, which then generates the policy gradient to update the  ``actor" network.
In AC, the key step is the accurate approximation of the value function.
Typically, there are three kinds of value functions. (1) State value: measure the performance of the current state, \ie, good or bad; (2) State-Action value (Q value): measure the performance of the determined action in the current state; (3) Advantage value: measure the advantage of the determined action than the average performance in the current state.

However, all of them are difficult to optimize PPN.
First, the state value is unsuitable for our case, since the action of PPN does not affect the environment state due to the general assumption (ii).
Thus, it cannot accurately measure the policy performance.
Next, the Q value is also unsuitable, since the Q network is often hard to train regarding the non-stationary decision process~\cite{moody2001learning}.
Finally, the advantage value is still inappropriate, since its optimization relies on the accurate estimations of both state and Q values.

In conclusion, the value functions are inappropriate for PPN, due to the difficult approximation  for portfolio selection.
Hence, they may lead to biased policy gradients and worse performance of AC.
On the contrary, DPG is guaranteed to obtain at least a sub-optimal solution as shown in Proposition~\ref{prop1}, and helps to obtain better performance.

We next empirically evaluate AC algorithms on Crypto-A.
We refer to the variant as PPN-AC, which is built upon Q values.
Specifically, we adopt the DDPG algorithm~\cite{Lillicrap2016} to optimize PPN-AC. The actor network in PPN-AC uses the same architecture as PPN, while the Q network and target Q network follow the network architecture in DDPG~\cite{Lillicrap2016}.
To better stabilize the training, we improve both Q networks with the dueling architecture mechanism~\cite{wang2016dueling}.

We record the detailed results on the Crypto-A dataset in Table~\ref{table7}. To be specific, the performance of PPN-AC is far worse than PPN.
This is because the Q network fails to approximate the Q value accurately, leading to biased policy gradients and worse performance.
Although PPN-AC cannot achieve a satisfactory result, it still performs better than other baselines in Table~\ref{table2}.
Such superiority mainly attributes to the strong representation ability of the actor network, \ie, PPN.
This further confirms the effectiveness of the two-stream architecture.
In the future, we will continue to improve the task-specific deep reinforcement learning algorithm for portfolio selection.

\begin{table}[h]
\caption{Evaluations of reinforcement learning algorithms for portfolio policy network on the Crypto-A dataset}
\vspace{-0.1in}
 \label{table7}
 \begin{center}
 \begin{small}
 \scalebox{1}{
	\begin{tabular}{|c|c|c|c|c|c|}\hline
		Algos& APV &STD($\small{\%}$) & SR($\small{\%}$) & MMD($\small{\%}$) & CR \\\hline \hline
          PPN-AC &11.72 & 2.73 &4.60& 60.25&17.79 \\
          PPN & 32.04  & 2.16  &6.85  & 38.86   &79.87 \\
          \hline
	\end{tabular}}
 \end{small}
 \end{center}
 \vspace{-0.35in}
\end{table}

\blue{\subsection{Application to Stock Portfolio Selection}
In  previous experiments, we have demonstrated the effectiveness of the proposed methods on  crypto-currency  datasets. Here, we further evaluate our methods on the  S$\&$P500 stock dataset obtained from Kaggle\footnote{https://www.kaggle.com/camnugent/sandp500}, which is summarized  in Table~\ref{data_sp}. All experimental settings and implementation details are the same as before, except that the results are averaged over 20 runs with random initialization seeds. The results in Table~\ref{table_new} further verify the effectiveness of the proposed method in terms of the superiority of reinforcement learning,  and the importance of sequential feature learning and  asset correlation extraction.  Also, the results demonstrate the generalization ability of our method.}

\begin{table}[h] 
\caption{\blue{The statistics of the S$\&$P500 dataset}} 
 \label{data_sp}
 \begin{center}
 \vspace{-0.1in}
 \begin{scriptsize}
 \scalebox{0.95}{
 \renewcommand*\arraystretch{1}
  \begin{tabular}{|c|c|c|c|c|c|}\hline
  \multirow{2}{*}{\hspace{-0.01in}Datasets\hspace{-0.01in}} & \multirow{2}{*}{\hspace{-0.047in} $\#$Asset \hspace{-0.047in}}&\multicolumn{2}{c|}{Training Data}&\multicolumn{2}{c|}{Testing Data}\cr
  \cline{3-6}
  & & Data Range& \hspace{-0.065in} Num. \hspace{-0.065in} & Data Range& \hspace{-0.06in} Num. \hspace{-0.06in} \cr
  \hline
       S$\&$P500  &  506 & 2013-02 to 2017-08   & 1101 &   2017-08 to 2018-02 	&  94  \\

  \hline%
  \end{tabular}}
  \end{scriptsize}
 \end{center}\vspace{-0.25in}
\end{table}

\section{Conclusion}
This paper has proposed a novel cost-sensitive portfolio policy network to solve the financial portfolio selection task.
Specifically, by devising a new two-stream architecture, the proposed network is able to extract both price sequential patterns and asset correlations.
In addition, to maximize the accumulated return while controlling both transaction and risk costs, we develop a new cost-sensitive reward function and adopt the direct policy gradient algorithm to optimize it.
We theoretically analyze the near-optimality of the reward  and show that the growth rate of the policy regarding this reward function can approach the theoretical optimum. 
We also empirically study the proposed method on real-world crypto-currency \blue{and stock} datasets.
Extensive experiments demonstrate its superiority in terms of profitability, cost-sensitivity and representation abilities.
In the future, we will further discuss two general assumptions, and continue to improve the task-specific deep reinforcement learning method for better effectiveness, stability and interpretability. \blue{For example, we can further explore the correlation between social text information and price sequential information~\cite{wu2018}.}

\section{Acknowledgement}
This work was partially supported by National Natural Science Foundation of China (NSFC) (No. 61876208), key project of NSFC (No. 61836003), Program for Guangdong Introducing Innovative and Enterpreneurial Teams 2017ZT07X183, Guangdong Provincial Scientific and Technological Funds (2018B010107001, 2018B010108002), Pearl River S$\&$T Nova Program of Guangzhou 201806010081, Tencent AI Lab Rhino-Bird Focused Research Program (No. JR201902),  Fundamental Research Funds for the Central Universities D2191240.

\begin{IEEEbiography}[{\includegraphics[width=1.2in,height=1.25in,keepaspectratio]{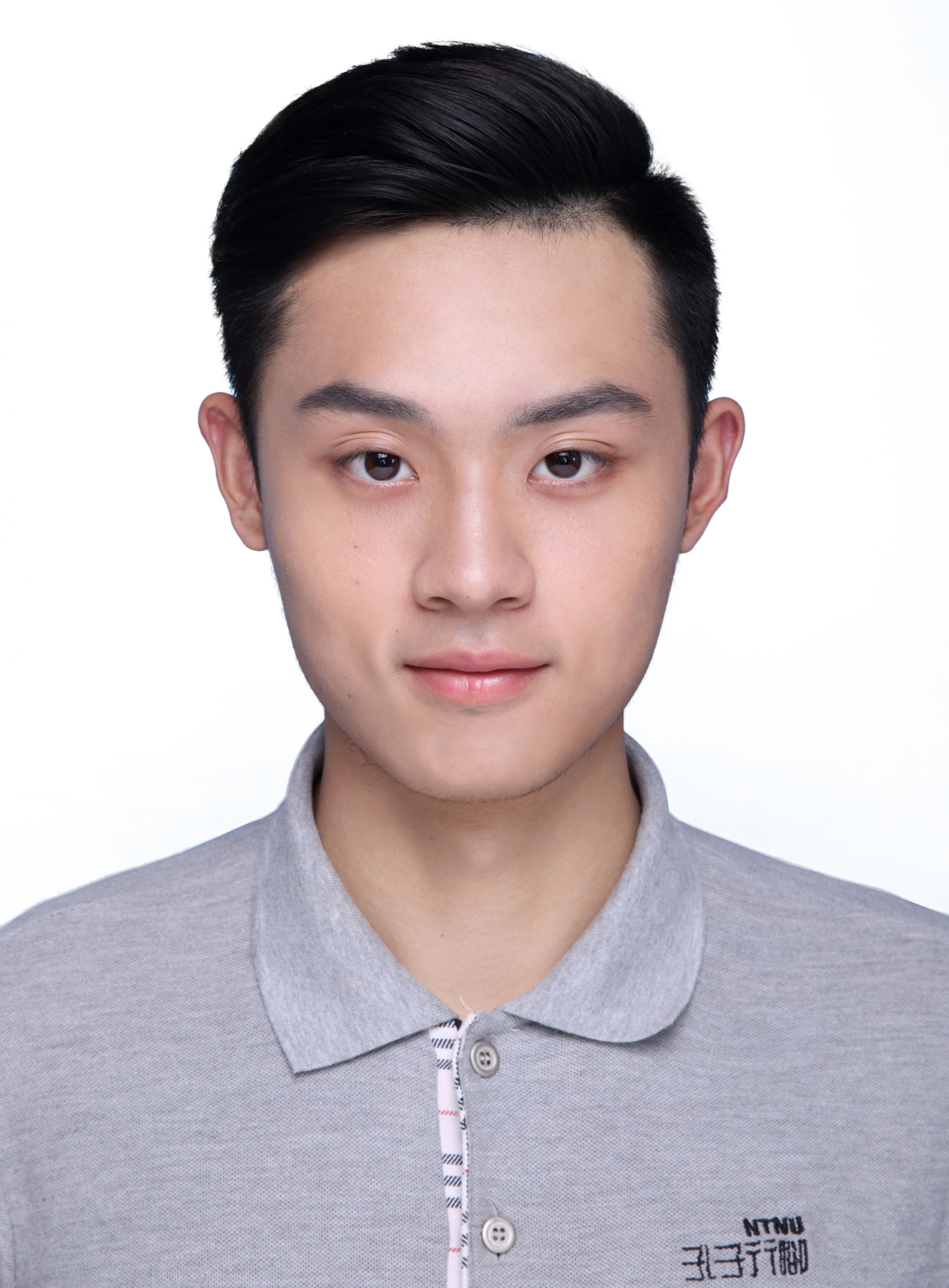}}]{Yifan Zhang}
is working toward the M.E. degree in the School of Software Engineering, South China University of Technology, China. He received the B.E. degree from the Southwest University, China, in 2017. His research interests are broadly in machine learning and data mining.  He has published papers in top venues including SIGKDD, NeurIPS, MICCAI, TKDE, and etc.  He has been invited as a PC member or reviewer for international conferences and journals, such as MICCAI and NeuroComputing.
\end{IEEEbiography}
\begin{IEEEbiography}[{\includegraphics[width=1.2in,height=1.25in,clip,keepaspectratio]{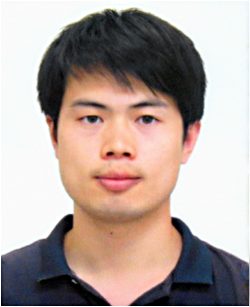}}]{Peilin Zhao}
is currently a Principal Researcher at Tencent AI Lab, China. Previously, he has worked at Rutgers University, Institute for Infocomm Research (I2R), Ant Financial Services Group. His research interests include: Online Learning, Recommendation System, Automatic Machine Learning, Deep Graph Learning, and Reinforcement Learning etc. He has published over 100 papers in top venues, including JMLR, ICML, KDD, etc. He has been invited as a PC member, reviewer or editor for many international conferences and journals, such as ICML, JMLR, etc. He received his bachelor’s degree from Zhejiang University, and his Ph.D. degree from Nanyang Technological University.
\end{IEEEbiography}

\begin{IEEEbiography}[{\includegraphics[width=1.2in,height=1.2in,clip,keepaspectratio]{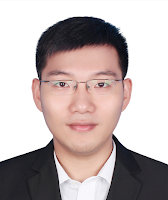}}]{Qingyao Wu} is
currently a Professor with the School
of Software Engineering, South China University
of Technology, Guangzhou, China. He
received the Ph.D. degree in computer
science from the Harbin Institute of Technology,
Harbin, China, in 2013. He was a Post-Doctoral Research Fellow with the School of
Computer Engineering, Nanyang Technological
University, Singapore, from 2014 to 2015.  His current
research interests include machine learning,
data mining, big data research.
\end{IEEEbiography}

\begin{IEEEbiography}[{\includegraphics[width=1.2in,height=1.25in,clip,keepaspectratio]{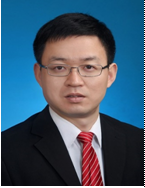}}]{Bin Li}
is currently a professor in
the Department of Finance, Economics and Management School at Wuhan University, Wuhan,
China.
He received the bachelor’s degree in computer
science from the Huazhong University of Science
and Technology and the bachelor’s degree in
economics from Wuhan University in 2006, and
the PhD degree from the School of Computer
Engineering at Nanyang Technological University
in 2013.  He was a postdoctoral research staff in the
Nanyang Business School at Nanyang Technological University, Singapore. His research interests are quantitative
investment, computational finance, and machine learning.
\end{IEEEbiography}

\begin{IEEEbiography}[{\includegraphics[width=1in,height=1.25in, keepaspectratio]{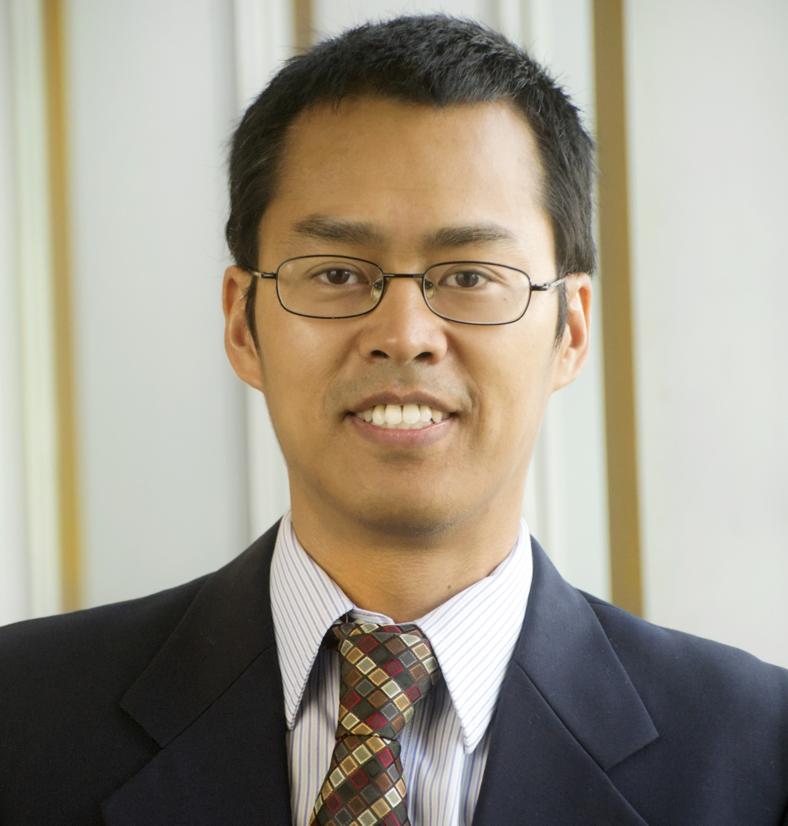}}]{Junzhou Huang}
is an Associate Professor in the Computer Science and Engineering department at the University of Texas at Arlington. He received the B.E. degree from Huazhong University of Science and Technology, China, the M.S. degree from Chinese Academy of Sciences, China, and the Ph.D. degree in Rutgers university. His major research interests include machine learning, computer vision and imaging informatics. He was selected as one of the 10 emerging leaders in multimedia and signal processing by the IBM T.J. Watson Research Center in 2010. He received the NSF CAREER Award in 2016.
\end{IEEEbiography}
 
\begin{IEEEbiography}[{\includegraphics[width=1.2in,height=1.25in,keepaspectratio]{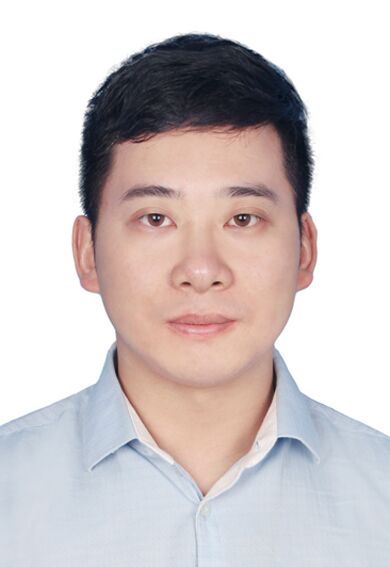}}]{Mingkui Tan}
 is currently a professor with the School of Software Engineering at South China University of Technology. He received his Bachelor Degree in Environmental Science and Engineering in 2006 and Master degree in Control Science and Engineering in 2009, both from Hunan University in Changsha, China. He received the Ph.D. degree in Computer Science from Nanyang Technological University, Singapore, in 2014. From 2014-2016, he worked as a Senior Research Associate on computer vision in the School of Computer Science, University of Adelaide, Australia. His research interests include machine learning, sparse analysis, deep learning and large-scale optimization.
\end{IEEEbiography}
 
\end{document}


\title{Cost-Sensitive Portfolio Selection via Deep Reinforcement Learning}

%


\author{Yifan Zhang, Peilin Zhao, Qingyao Wu, Bin Li, Junzhou Huang, and Mingkui Tan 
\IEEEcompsocitemizethanks{
\IEEEcompsocthanksitem Y. Zhang, Q. Wu and M. Tan are with South China University of Technology and Guangzhou  Laboratory, China. E-mail: sezyifan@mail.scut.edu.cn; $\{$qyw, mingkuitan$\}$@scut.edu.cn. 
\IEEEcompsocthanksitem P. Zhao and J. Huang are with Tencent AI Lab, China. Email: peilinzhao@hotmail.com; joehhuang@tencent.com.
\IEEEcompsocthanksitem B. Li is with Wuhan University, China. E-mail: binli.whu@whu.edu.cn. 
\IEEEcompsocthanksitem P. Zhao, Q. Wu are co-first authors. M. Tan is the corresponding author.}
} 


\markboth{IEEE TRANSACTIONS ON KNOWLEDGE AND DATA ENGINEERING}%
{Zhang \MakeLowercase{\textit{et al.}}}

\IEEEtitleabstractindextext{
\begin{abstract}
This supplemental file provides the proofs of theorems \blue{and more detailed empirical results} in our paper of "Cost-Sensitive Portfolio Selection via Deep Reinforcement Learning"~\cite{zhang2018online}.
\end{abstract}

\begin{IEEEkeywords}
Portfolio Selection; Reinforcement Learning; Deep Learning; Transaction Cost.
\end{IEEEkeywords}}

\maketitle

\IEEEdisplaynontitleabstractindextext

\IEEEpeerreviewmaketitle

\appendices
\IEEEraisesectionheading{\section{Proofs of Theorems 1}\label{theorems}}

\begin{theorem}
Let $\bar{W}^*$ be the growth rate of the log-optimal strategy and $S^*_t$ be the wealth achieved by the optimal portfolio policy that maximizes $\mathbb{E}\{\log r_t\} \small{-} \lambda Var\{\log r_t\}$. If there is no transaction cost and the market is stationary ergodic, for any $\lambda\small{\geq} 0$ and $\frac{1}{e}\small{\leq} r_t \small{\leq} e$, the maximal growth rate of this policy satisfies:
\begin{align}
  \bar{W}^* \geq  \liminf_{t\small{\rightarrow}\infty}\frac{1}{t}\log S^*_t \geq \bar{W}^*-\frac{9}{4}\lambda. \nonumber
\end{align}
\end{theorem}

{\bf Proof.} To prove Theorem 1, let us introduce some notations. First, all price vectors $x_1,x_2,...$ are realizations of the random vectors $X_1,X_2,...$ drawn from the vector-valued stationary and ergodic process $\{X\}_{-\infty}^{\infty}$. Based on the Markov property, we are able to make the decision according to the current state:
\begin{align}
a_t=a(\{X\}_1^{t-1})=a(X_{t-1}). \nonumber
\end{align}

After the decision, the immediate reward on the $t$-th period is $r_t \small{=}a_t^{\top} X_t$. We then define the optimal portfolio policy that maximizing $\mathbb{E}\{\log r_t\} \small{-} \lambda Var\{\log r_t\}$ as $A^*\small{=} {\{a^*}\}$, where
\begin{align}
  a^*_t\small{:=}\text{argmax}_{a} \mathbb{E}\{\log r_t|X_{t-1}\} \small{-} \lambda Var\{\log r_t|X_{t-1}\}. \nonumber
\end{align}

In addition, we define the log-optimal  policy that maximizes the expected log-return $\mathbb{E}\{\log r_t\}$ as $ \bar{A}^*\small{=} {\{\bar{a}^*}\}$, where
\begin{align}
  \bar{a}^*_t\small{:=} \text{argmax}_{\bar{a}} \mathbb{E}\{\log \bar{a}_t^{\top}X_t|X_{t-1}\}. \nonumber
\end{align}

To simplifying the notations, we denote the conditional expected value of the log-return as:
\begin{align}
  E(\log r_t):= \mathbb{E}\{\log r_t|X_{t-1}\}, \nonumber
\end{align}
and denote the conditional second order moment of the log-return as:
\begin{align}
  E(\log r_t)^2:= \mathbb{E}\{(\log r_t)^2|X_{t-1}\}, \nonumber
\end{align}
and finally denote the conditional variance of the log-return as:
\begin{align}
  Var(\log r_t):= E(\log r_t)^2 - E^2(\log r_t). \nonumber
\end{align}

Based on these notations, we redefine the reward function  $\mathbb{E}\{\log r_t\} \small{-} \lambda Var\{\log r_t\}$ as:
\begin{align}\label{ap1}
 h(r_t):=E(\log r_t)-\lambda[E(\log r_t)^2 - E^2(\log r_t)].
\end{align}

Next, we need to find the relationship between $\mathbb{E}\{\log r_t\}$ and $h(r_t)$ under the condition $\frac{1}{e}\leq r_t\leq e$.

First, when $r_t\in [\frac{1}{e},1]$, we have:
\begin{align}
-1\leq E(\log r_t)\leq 0, \nonumber
\end{align}
since the expected value of a variable is also bounded by its maximal/minimal values.

Then, we can easily obtain:
\begin{align}
  E(\log r_t)\leq E(\log r_t)^2 , \nonumber
\end{align}
and thus have:
\begin{align}
  E(\log r_t)- E^2(\log r_t)\leq E(\log r_t)^2 - E^2(\log r_t). \nonumber
\end{align}

Since the left-hand term is a quadratic function, and $-1\leq E(\log r_t)\leq 0$, we easily obtain:
\begin{align}
  -2 \leq E(\log r_t)- E^2(\log r_t) \leq 0, \nonumber
\end{align}
and thus have:
\begin{align}\label{ap2}
  E(\log r_t)^2 - E^2(\log r_t) \geq -2.
\end{align}

In addition, when $r_t\in (1,e]$, we have:
\begin{align}
  E(\log r_t)\geq E(\log r_t)^2, \nonumber
\end{align}
and thus:
\begin{align}
  E(\log r_t)- E^2(\log r_t) \geq E(\log r_t)^2 - E^2(\log r_t). \nonumber
\end{align}

Since the left-hand term is a quadratic function, and $r_t\in (1,e]$, \ie $0< E(\log r_t)\leq 1$, we can easily obtain:
\begin{align}
  0 \leq E(\log r_t)- E^2(\log r_t) \leq \frac{1}{4}, \nonumber
\end{align}
and thus have:
\begin{align}\label{ap3}
  E(\log r_t)^2 - E^2(\log r_t) \leq \frac{1}{4}.
\end{align}

Since $E(\log r_t)^2 - E^2(\log r_t)$ is a discrete value, by combining Eqn.~(\ref{ap2}) and Eqn.~(\ref{ap3}), we can constrain $E(\log r_t)^2 - E^2(\log r_t)$ by a loose bound:
\begin{align}
 -\frac{1}{4}\lambda \leq -\lambda [E(\log r_t)^2 - E^2(\log r_t)] \leq 2\lambda. \nonumber
\end{align}

Then, based on Eqn.~(\ref{ap1}), we have:
\begin{align}
 E(\log r_t) -\frac{1}{4}\lambda \leq h(r_t) \leq E(\log r_t) +2\lambda. \nonumber
\end{align}

By taking the definitions of both optimal policies that optimizing different reward functions, we have:
\begin{align}
  & E(\log a_t^{* \top}X_t)+ 2\lambda \nonumber \\
  \geq \ \ \ & h( a_t^{* \top}X_t) \nonumber \\
  \geq \ \ \ & h( \bar{a}_t^{* \top}X_t)\nonumber \\
  \geq \ \ \ & E(\log \bar{a}_t^{* \top}X_t)- \frac{1}{4}\lambda. \nonumber
\end{align}

Thus, we obtain:
\begin{align}\label{ap4}
  E(\log a_t^{* \top}X_t) \geq E(\log \bar{a}_t^{* \top}X_t)- \frac{9}{4}\lambda.
\end{align}

Following \cite{gyorfi2008growth,ottucsak2007asymptotic}, consider the following decomposition
\begin{align}
  \frac{1}{t}\log S_t^*=U_t^* + V_t^*, \nonumber
\end{align}
where
\begin{align}
  U_t^* =\frac{1}{t}\sum_{i=1}^t\big{[}\log a_i^{*\top}X_i-E(\log a_i^{*\top}X_i)\big{]}, \nonumber
\end{align}
and
\begin{align}
  V_t^* =\frac{1}{t}\sum_{i=1}^t\big{[}E(\log a_i^{*\top}X_i)\big{]}. \nonumber
\end{align}

Here, $U_t^{\ast}$ is an average of bounded martingale differences. According to the Chow Theorem (cf. Theorem 3.3.1 in \cite{stout1974almost}), $U_t^{\ast}$ converges to 0 almost surely, since $ \sum_{i=1}^{\infty}\frac{E\{(\log a_i^{*\top}X_i)^2\}}{i^2}<\infty$ implies that $U_t^{\ast} \rightarrow 0$ almost surely. So,
\begin{align}
  \liminf_{n\rightarrow \infty}\frac{1}{t}\log S_t^*=\liminf_{n\rightarrow \infty}V_t^*. \nonumber
\end{align}

Next, let $\bar{S}_t^*$ be the wealth achieved by the log-optimal portfolio strategy. Similarly, consider the following decomposition
\begin{align}
  \frac{1}{t}\log \bar{S}_t^*=\bar{U}_t^* + \bar{V}_t^*, \nonumber
\end{align}
where
\begin{align}
  \bar{U}_t^* =\frac{1}{t}\sum_{i=1}^t\big{[}\log \bar{a}_i^{*\top}X_i-E(\log \bar{a}_i^{*\top}X_i)\big{]}, \nonumber
\end{align}
and
\begin{align}
  \bar{V}_t^* =\frac{1}{t}\sum_{i=1}^t\big{[}E(\log \bar{a}_i^{*\top}X_i)\big{]}. \nonumber
\end{align}

Again, it can be showed that $\bar{U}_t^{\ast} \rightarrow 0$. Hence
\begin{align}
  \lim_{n\rightarrow \infty}\frac{1}{t}\log \bar{S}_t^*=\lim_{n\rightarrow \infty}\bar{V}_t^*. \nonumber
\end{align}

Taking the limes inferior of both sides as $n$ goes to infinity and taking the arithmetic average on both sides of Eqn.~(\ref{ap4}) over trading periods $1,...,t$, we obtain the theoretical maximal growth rate regarding the risk-sensitive reward:
\begin{align}
  \bar{W}^* \geq \liminf_{n\rightarrow \infty}\frac{1}{t}\log S_t^* \geq \bar{W}^*-\frac{9}{4}\lambda, \nonumber
\end{align}
where $\bar{W}^*\small{=}\lim_{t\small{\rightarrow} \infty}\frac{1}{t}\log \bar{S}^*_t$ is the  growth rate of the log-optimal portfolio strategy.

Thus, we conclude the proofs of Theorem 1.

\section{Proofs of Proposition 4}
\begin{theorem}
Let $\psi$ be the transaction cost rate, $\hat{a}_{t-1}$ and $a_t$ be the asset allocations before and after rebalancing, then the cost proportion $c_t$ on the $t$-th period is bounded:
\begin{align}
  \frac{\psi}{1+\psi}\|a_t\small{-}\hat{a}_{t-1}\|_1 \leq c_t \leq \frac{\psi}{1-\psi}\|a_t\small{-}\hat{a}_{t-1}\|_1, \nonumber
\end{align}
where $\|a_t\small{-}\hat{a}_{t-1}\|_1 \in \Big{(}0,\frac{2(1-\psi)}{1+\psi}\Big{]}$.
\end{theorem}

{\bf Proof.} To bound the cost proportion $c_t$, we need to get rid of the $w_t$ within the L1 norm $c_t = \psi \|a_t\omega_{t} -\hat{a}_{t-1}\|_1$ in the main text. By using the norm inequality, we have:
\begin{align}
  c_t &= \psi \|a_t\omega_{t} -\hat{a}_{t-1}\|_1 \nonumber \\ \nonumber
  &= \psi \|a_t -a_t c_t -\hat{a}_{t-1}\|_1 \\ \nonumber
  &\leq \psi \|a_t -\hat{a}_{t-1}\|_1 + \psi c_t \|a_t\|_1 \\ \nonumber
  &= \psi \|a_t -\hat{a}_{t-1}\|_1 + \psi c_t.
\end{align}

The upper bound of $c_t$ is:
\begin{align}
  c_t \leq \frac{\psi}{1-\psi}\|a_t\small{-}\hat{a}_{t-1}\|_1. \nonumber
\end{align}

Similarly, we can obtain the lower bound as follows:
\begin{align}
  c_t &= \psi \|a_t -a_t c_t -\hat{a}_{t-1}\|_1 \nonumber \\ \nonumber
  &\geq \psi \|a_t -\hat{a}_{t-1}\|_1 - \psi c_t \|a_t\|_1 \\ \nonumber
  &= \psi \|a_t -\hat{a}_{t-1}\|_1 - \psi c_t.
\end{align}

Thus, the lower bound of $c_t$ is:
\begin{align}
  c_t \geq \frac{\psi}{1+\psi}\|a_t\small{-}\hat{a}_{t-1}\|_1. \nonumber
\end{align}

In conclusion, the upper/lower bounds of $c_t$ are:
\begin{align}\label{ap5}
 \frac{\psi}{1+\psi}\|a_t\small{-}\hat{a}_{t-1}\|_1 \leq c_t \leq \frac{\psi}{1-\psi}\|a_t\small{-}\hat{a}_{t-1}\|_1 .
\end{align}

We then constrain the L1 norm $\|a_t\small{-}\hat{a}_{t-1}\|_1$. Let $N_t$ be the net wealth of portfolios and $W_t$ be the gross wealth at the end of market dat $t$, where the net proportion $\omega_t = \frac{N_t}{W_t}$ and $W_t = N_{t-1}a_t x_t$.

Following \cite{ormos2013performance}, the income from the sales decreased by the transaction costs of sales, covers the value of purchases and the transaction costs of purchases. That is:
\begin{align}
  &\sum_{i=1}^m \Big{\{} \big{(}a_{t-1,i}x_{t-1,i}N_{t-1} - a_{t,i}N_t\big{)}^{+} \nonumber \\
  & \hspace{0.7in}- \psi_s \big{(}a_{t-1,i}x_{t-1,i}N_{t-1} - a_{t,i}N_t\big{)}^{+}\Big{\}} \nonumber \\
  =&\sum_{i=1}^m \Big{\{} \big{(}a_{t,i}N_t-a_{t-1,i}x_{t-1,i}N_{t-1}\big{)}^{+} \nonumber \\
  &\hspace{0.7in}+ \psi_p \big{(}a_{t,i}N_t-a_{t-1,i}x_{t-1,i}N_{t-1}\big{)}^{+}\Big{\}}.\nonumber
\end{align}
This equals to:
\begin{align}
  &(1-\psi_s)\sum_{i=1}^m \big{(}a_{t-1,i}x_{t-1,i}N_{t-1} - a_{t,i}N_t\big{)}^{+} \nonumber \\
  =& (1+\psi_p) \sum_{i=1}^m \big{(}a_{t,i}N_t-a_{t-1,i}x_{t-1,i}N_{t-1}\big{)}^{+}.\nonumber
\end{align}
Then, using the identity
\begin{align}
  (a-b)^+=a-b+(b-a)^+, \nonumber
\end{align}
we have:
\begin{align}
  &(1-\psi_s)\sum_{i=1}^m \big{(}a_{t-1,i}x_{t-1,i}N_{t-1} - a_{t,i}N_t\big{)}^{+} \nonumber \\
  =& (1+\psi_p) \bigg{\{}\sum_{i=1}^m \big{(}a_{t,i}N_t-a_{t-1,i}x_{t-1,i}N_{t-1}\big{)}  \nonumber \\
  &\hspace{1in} + \sum_{i=1}^m \big{(}a_{t-1,i}x_{t-1,i}N_{t-1} - a_{t,i}N_t\big{)}^{+}\bigg{\}}.\nonumber
\end{align}
Since $\sum_{i=1}^m \big{(} a_{t,i}N_t\big{)} \small{=}N_{t}$ and $\sum_{i=1}^m \big{(} a_{t-1,i}x_{t-1,i}N_{t-1}\big{)} =W_{t}$, we obtain:
\begin{align}
  W_t - N_t = \frac{\psi_p+\psi_s}{1+\psi_p}\sum_{i=1}^m \big{(}a_{t-1,i}x_{t-1,i}N_{t-1} - a_{t,i}N_t\big{)}^{+}. \nonumber
\end{align}
Dividing both sides by $W_t$, we have:
\begin{align}
  1 - \omega_t= \frac{\psi_p+\psi_s}{1+\psi_p}\sum_{i=1}^m \big{(}\hat{a}_{t-1,i} - a_{t,i}\omega_t\big{)}^{+}, \nonumber
\end{align}
where $\hat{a}_{t-1,i}\small{=}\frac{a_{t-1}\odot x_{t-1}}{a_{t-1}x_{t-1}}$. This equation equals to:
\begin{align}
  c_t= \frac{\psi_p+\psi_s}{1+\psi_p}\sum_{i=1}^m \big{(}\hat{a}_{t-1,i} - a_{t,i}\omega_t\big{)}^{+}. \nonumber
\end{align}

According to the definition of proportional transaction cost, and adjusting the above equation with $\psi_p=\psi_s=\psi$ , we have the following tighter bound:
\begin{align}
 0<c_t\leq \frac{2\psi}{1+\psi}. \nonumber
\end{align}
Combining the derived bounds of $c_t$ in Eqn.~(\ref{ap5}),  $\|a_t -\hat{a}_{t-1}\|_1$ is bounded as:
\begin{align}
 0< \|a_t -\hat{a}_{t-1}\|_1 \leq \frac{2(1-\psi)}{1+\psi}, \nonumber
\end{align}
which concludes the proofs of Theorem 2.

\section{Proofs of Theorem 3}
\begin{theorem}
Let $\tilde{W}^*$ be the growth rate of the theoretical optimal strategy that optimizes $\mathbb{E}\{\log r^c_t\}$, and $S^*_t$ be the wealth achieved by the optimal portfolio policy that maximizes $\mathbb{E}\{\log r_t^c\} \small{-} \lambda Var\{\log r_t^c\} \small{-} \gamma \mathbb{E}\{\|a_t\small{-}\hat{a}_{t-1}\|_1\}$. If the market is stationary and the return process is a homogeneous and first-order Markov process, for any $\lambda\geq 0$, $\gamma\geq0$, $\psi \in[0,1]$ and $\frac{1}{e}\small{\leq} r_t^c \small{\leq} e$, the maximal growth rate of this policy satisfies:
\begin{align}
  \tilde{W}^* \geq  \liminf_{t\small{\rightarrow}\infty}\frac{1}{t}\log S^*_t > \tilde{W}^*-\frac{9}{4}\lambda - \frac{2\gamma(1-\psi)}{1+\psi}. \nonumber
\end{align}
\end{theorem}


{\bf Proof.} To simplify the formulation, we keep using the notations in the proof of Theorem 1. For clarity, we present them again.
All price vectors $x_1,x_2,...$ are realizations of the random vectors $X_1,X_2,...$ drawn from the vector-valued stationary process $\{X\}_{-\infty}^{\infty}$.

Based on the Markov property, we make the decision by:
\begin{align}
a_t=a(\{X\}_1^{t-1})=a(X_{t-1}). \nonumber
\end{align}
And the immediate cost-adjusted reward on the $t$-th period is $r_t^c \small{=}(1-c_t)*a_t^{\top} X_t$.

We define the optimal portfolio policy that maximizing the cost-sensitive reward, as $A^*\small{=} {\{a^*}\}$, where
\begin{align}
  a^*_t\small{:=}\text{argmax}_{a} \mathbb{E}\{\log r_t^c|X_{t-1}\} &\small{-} \lambda Var\{\log r_t^c|X_{t-1}\}  \nonumber \\
  & \small{-} \gamma \mathbb{E}\{\|a_t\small{-}\hat{a}_{t-1}\|_1|X_{t-1}\}. \nonumber
\end{align}

We next define the optimal portfolio policy that maximizing $E(\log r_t^c)$, as $ \tilde{A}^*\small{=} {\{\tilde{a}^*}\}$, where
\begin{align}
  \tilde{a}^*_t\small{:=}\text{argmax}_{\tilde{a}} \mathbb{E}\{\log \big{(}(1-\tilde{c}_t)*\tilde{a}_t^{\top}X_t \big{)}|X_{t-1}\}, \nonumber
\end{align}
where $\tilde{c}_t$ is the proportional transaction cost of this optimal policy.

Similarly, we denote the conditional expected value of the cost-adjusted log-return as:
\begin{align}
  E(\log r_t^c)= \mathbb{E}\{\log r_t^c|X_{t-1}\}, \nonumber
\end{align}
and denote the conditional second order moment as:
\begin{align}
  E(\log r_t^c)^2= \mathbb{E}\{(\log r_t^c)^2|X_{t-1}\}, \nonumber
\end{align}
and finally denote the conditional variance as:
\begin{align}
  Var(\log r_t^c)= E(\log r_t^c)^2 - E^2(\log r_t^c). \nonumber
\end{align}

Based on these notations, we redefine the cost-sensitive reward function as:
\begin{align}
 z(r_t^c):=E(\log r_t^c) &\small{-}\lambda[E(\log r_t^c)^2 \small{-} E^2(\log r_t)] \nonumber \\
 & - \gamma \mathbb{E}\{\|a_t-\hat{a}_{t-1}\|_1|X_{t-1}\}. \nonumber
\end{align}


According to Theorem 2, we have
\begin{align}
\|a_t\small{-}\hat{a}_{t-1}\|_1 \in \Big{(}0,\frac{2(1-\psi)}{1+\psi}\Big{]}.  \nonumber
\end{align}

Since the expected value of a variable is also bounded by its maximal/minimal values, we thus have:
\begin{align}
 0 < \mathbb{E}\{\|a_t\small{-}\hat{a}_{t-1}\|_1|X_{t-1}\} \leq \frac{2(1-\psi)}{1+\psi}. \nonumber
\end{align}

Combining with the results in Theorem 1, we have:
\begin{align}
  & E \big{(}\log \big{(}(1-c_t)*a_t^{\top}X_t \big{)}\big{)}+ 2\lambda \nonumber \\
  > \ \ \ & z \big{(}(1-c_t)*a_t^{* \top}X_t\big{)} \nonumber \\
  \geq \ \ \ & z \big{(}(1-\tilde{c}_t)*\tilde{a}_t^{* \top}X_t\big{)}\nonumber \\
  \geq \ \ \ & E\big{(}\log \big{(}(1-\tilde{c}_t)*\tilde{a}_t^{\top}X_t \big{)}\big{)}- \frac{1}{4}\lambda -\frac{2\gamma(1-\psi)}{1+\psi}. \nonumber
\end{align}

We thus obtain:
\begin{align}\label{ap6}
  & E \big{(}\log \big{(}(1-c_t)*a_t^{\top}X_t \big{)}\big{)}  \nonumber \\
  > & E\big{(}\log \big{(}(1-\tilde{c}_t)*\tilde{a}_t^{\top}X_t \big{)}\big{)}- \frac{9}{4}\lambda -\frac{2\gamma(1-\psi)}{1+\psi}.
\end{align}

As Consider the following decomposition
\begin{align}
  \frac{1}{t}\log S_t^*=U_t^* + V_t^*, \nonumber
\end{align}
where
\begin{align}
  U_t^* \small{=}\frac{1}{t}\sum_{i=1}^t\big{[}\log ((1\small{-}c_i)\small{*}a_i^{*\top}X_i )\small{-}E \big{(}\log ((1\small{-}c_i)\small{*}a_i^{*\top}X_i )\big{)}\big{]}, \nonumber
\end{align}
and
\begin{align}
  V_t^* =\frac{1}{t}\sum_{i=1}^t\big{[}E \big{(}\log \big{(}(1\small{-}c_i)\small{*}a_i^{*\top}X_i \big{)}\big{)}\big{]}. \nonumber
\end{align}

Similar to the proofs of Theorem 1, $U_t^{\ast}$ converges to 0 almost surely, \ie $U_t^{\ast} \rightarrow 0$. Thus
\begin{align}
  \liminf_{n\rightarrow \infty}\frac{1}{t}\log S_t^* = \liminf_{n\rightarrow \infty}V_t^*. \nonumber
\end{align}

Then, let $\tilde{S}_t^*$ be the wealth achieved by the optimal portfolio strategy that maximizing $E(\log r_t^c)$. Similarly, consider the following decomposition
\begin{align}
  \frac{1}{t}\log \tilde{S}_t^*=\tilde{U}_t^* + \tilde{V}_t^*, \nonumber
\end{align}
where
\begin{align}
  \tilde{U}_t^* \small{=}\frac{1}{t}\sum_{i=1}^t\big{[}\log ((1\small{-}\tilde{c}_i)\small{*}\tilde{a}_i^{*\top}X_i )\small{-}E \big{(}\log ((1\small{-}\tilde{c}_i)\small{*}\tilde{a}_i^{*\top}X_i )\big{)}\big{]}, \nonumber
\end{align}
and
\begin{align}
  \tilde{V}_t^* =\frac{1}{t}\sum_{i=1}^t\big{[}E \big{(}\log \big{(}(1\small{-}\tilde{c}_i)\small{*}\tilde{a}_i^{*\top}X_i \big{)}\big{)}\big{]}. \nonumber
\end{align}

Again, it can be showed that $\tilde{U}_t \rightarrow 0$. Hence
\begin{align}
  \lim_{n\rightarrow \infty}\frac{1}{t}\log \tilde{S}_t^*= \lim_{n\rightarrow \infty}\tilde{V}_t^*. \nonumber
\end{align}

Taking the limes inferior of both sides as $n$ goes to infinity and taking the arithmetic average on both sides of Eqn.~(\ref{ap6}) over trading periods $1,...,t$, we obtain the theoretical maximal growth rate regarding the cost-sensitive reward:
\begin{align}
  \tilde{W}^* \geq \liminf_{n\rightarrow \infty}\frac{1}{t}\log S_t^*> \tilde{W}^*-\frac{9}{4}\lambda -\frac{2\gamma(1-\psi)}{1+\psi}, \nonumber
\end{align}
where $\tilde{W}^*\small{=}\lim_{t\small{\rightarrow} \infty}\frac{1}{t}\log \tilde{S}^*_t$ is the growth rate of the theoretical optimal strategy that optimizes $\mathbb{E}\{\log r^c_t\}$.

Thus, we conclude the proofs of Theorem 3.

\begin{figure*}[t]
  \vspace{0.1in}
  \begin{center}
  \includegraphics[width=17cm]{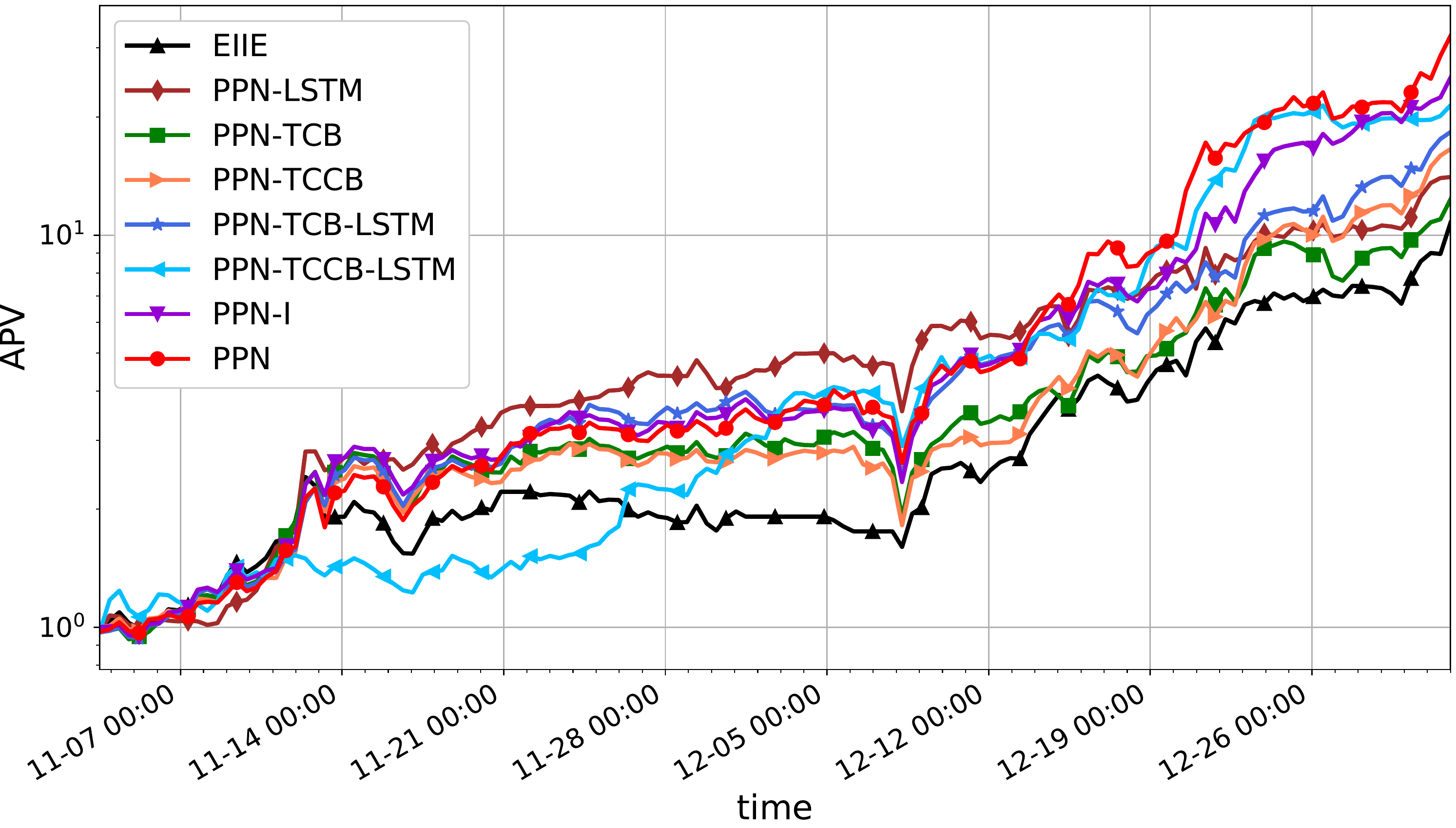}
  \end{center}
  \caption{\blue{The performance development of the proposed portfolio policy network with different feature extractors and EIIE on the Crypto-A dataset (Better viewed in color)}}
  \label{fig5}
  \vspace{-0.1in}
\end{figure*}

\begin{figure*}[h] 
  \centerline{\includegraphics[width=16cm]{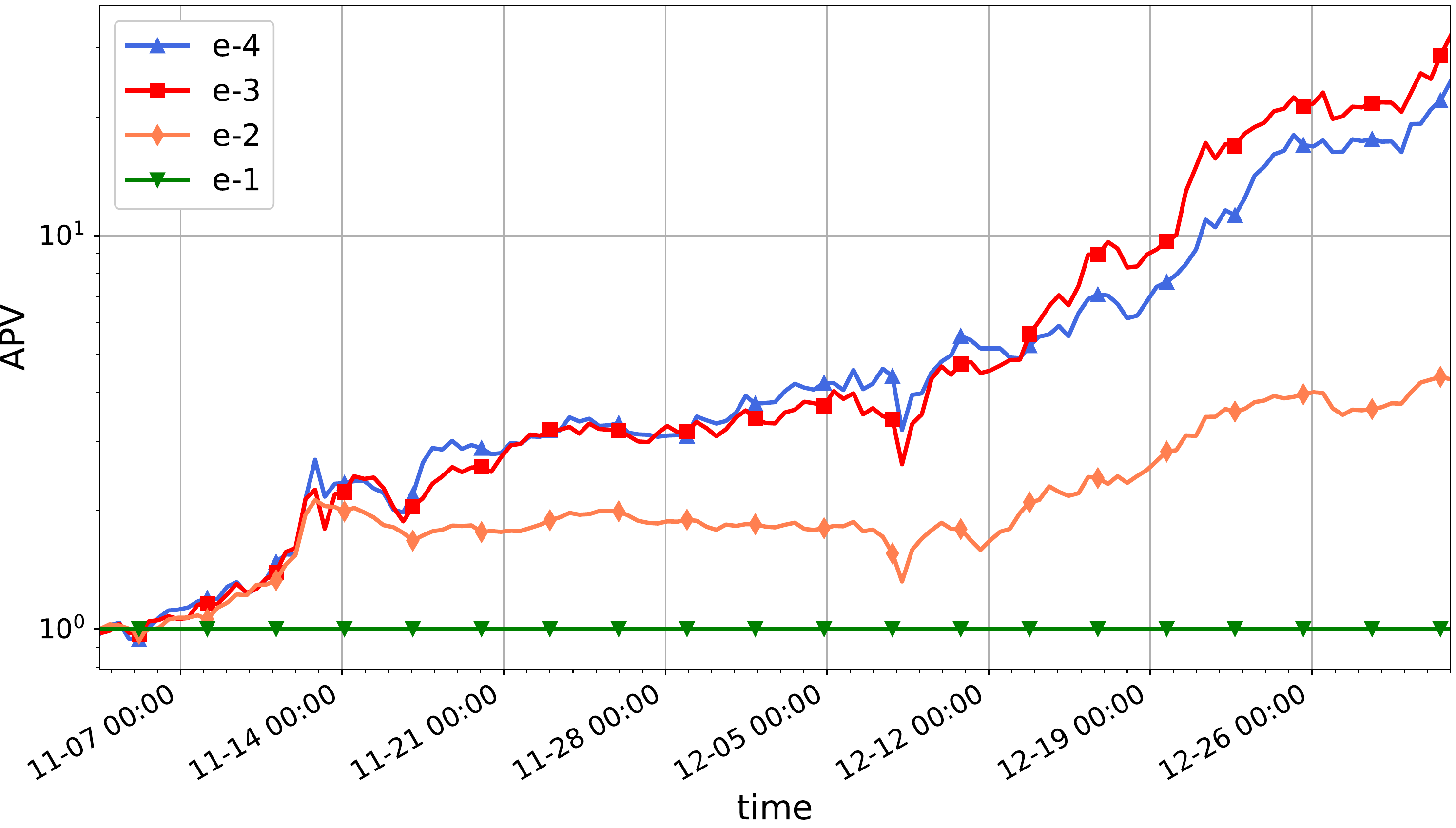}} 
  \caption{\blue{Performance development of portfolio policy network under different $\gamma$ on Crypto-A (Better viewed in color)}}
  \label{fig6}
 \vspace{-0.1in}
\end{figure*}

\blue{\section{More Experimental Results}
\subsection{Enlarged Figures}
In this subsection, we provide the larger scale of the Fig.~5 and Fig.~6 in the main text, as shown in Fig Sup.~\ref{fig5} and Fig Sup.~\ref{fig6}.}

\blue{\subsection{More Empirical Results}
In this subsection, we provide more detailed results of Tables 4-8 in the main text, as shown in Tables Sup.~\ref{table2}-\ref{table6}, while the detailed discussions can be found in the main text.}

\balance


\newpage
\begin{table*}[t]  
\caption{\blue{Performance comparisons on different datasets in terms of APV, SR($\small{\%}$), CR and TO}}
\vspace{-0.2in}
 \label{table2}
 \begin{center}
 \scalebox{0.87}{
	\begin{tabular}{|l|c|c|c|c|c|c|c|c|c|c|c|c|c|c|c|c|c|c|c|}\hline
  \multirow{2}{*}{Algos} & \multicolumn{4}{c|}{Crypto-A}&\multicolumn{4}{c|}{Crypto-B}&\multicolumn{4}{c|}{Crypto-C}&\multicolumn{4}{c|}{Crypto-D}\cr
  \cline{2-17}
  & APV & SR & CR & \blue{TO} & APV & SR&CR &  \blue{TO}  & APV & SR &CR &  \blue{TO}  & APV & SR &CR &  \blue{TO}  \cr
  \hline
      UBAH & 2.59 & 3.87 & 3.39  & 0.001        &1.63 & 2.57 & 2.43   & 0.001             & 1.32 & 3.00 & 1.61  &0.001 & 0.63 & 0.20 & -5.85 &0.001\\\hline
      
      Best &6.65 & 4.59 & 13.67   & 0.001          & 3.20 & 2.95 & 3.61      & 0.001         &2.97 & 3.15 & 3.96   &0.001 & 1.04 & 0.64 & 0.63 &0.001\\\hline
      
      CRP &2.40 & 3.95 & 3.24     & 0.009        &1.90 & 3.77 & 3.81  &0.010          & 1.30 & 3.30 & 1.77    &0.006   & 0.66 & 0.20 & -5.85 &0.006\\\hline
      
      UP & 2.43  & 3.95 & 3.28      & 0.009       &1.89 & 3.70 &3.71     &0.010   &1.30 & 3.29 & 1.76     &0.005     & 0.66 & 0.20 &  -5.85 &0.006\\\hline
      
      EG & 2.42 & 3.96 & 3.27     & 0.009         &1.89 & 3.71 & 3.75     &0.010        & 1.30 & 3.30 & 1.76  &0.005   & 0.67 & 0.21 &  -5.85 &0.006\\\hline
      
      Anticor &2.17 & 2.96 & 2.23      & 0.025      &21.80 & 9.92 & 103.68  & 0.348        & 0.75 & -1.48 & -0.91  &0.343   &  3.14 & 6.81 & 66.28& 0.432\\\hline
      
      ONS & 1.28  & 1.40 &  0.53   & 0.028            &1.71 & 3.15 & 4.00    & 0.040         & 1.14 & 2.33 & 1.95    &0.031   & 1.00 & 0.19 & 0.01 & 0.070\\\hline
      
      CWMR & 0.01 & -8.21 & -0.99  & 1.700    &0.64 & 0.42 & -0.54    &1.771        & 0.01 & -16.61 & -0.99    &1.814  & 0.38 & -0.02 &  -5.19 & 1.861\\\hline
      
      PAMR & 0.01 & -7.17 & -0.99   &1.646      &0.880 & 0.91 & -0.20       &1.736       & 0.01 & -15.48 & -0.99  &1.757 & 0.82 & 0.08 & -1.78 & 1.849\\\hline
      
      OLMAR &0.65 & 0.32 & -0.47      &1.270      &774.47 & 10.91 & 2040.70       &1.389      & 0.05 & -7.56 & -0.99 & 1.346      & 11.25 & 7.21 & 135.97 & 1.475\\\hline
      
      RMR & 0.69 & 0.46  & -0.42    & 1.228         & 842.26 & 11.62 & 3387.69 & 1.349         &0.05 & -7.72 & -0.99  & 1.334      & 14.337& 7.93 & 192.59 & 1.444\\\hline
      
      WMAMR & 0.85 & 0.67 &  -0.22  & 0.799        &87.85 & 8.25 & 245.23   & 0.913              & 0.26 & -3.78 & -0.98&0.809 & 7.72 & 6.62 &  227.16 & 0.961\\\hline
      
      EIIE &10.48  & 5.27 & 21.47  & 0.471       & 2866.15 & 13.42 & 8325.78   & 0.948        & 2.87 & 4.04 & 9.54  &0.359       & 113.58 & 15.11 & 4670.91 & 1.108\\\hline
      
      PPN-I & 25.76  & 6.75 & 57.05 &0.316       & 7549.35 &  14.74 & 28915.43 & 0.796   & 3.93 & 5.12 & 15.04 &0.331    &  238.93 & 16.07 & 8803.95 & 0.937\\\hline
      
      PPN & \textbf{32.04} & \textbf{6.85} &\textbf{79.87}  &0.368        &\textbf{9842.56} & \textbf{14.82}  &\textbf{37700.03}    & 0.888         &\textbf{4.81} & \textbf{5.89} & \textbf{16.11} &0.304  & \textbf{538.22} & \textbf{17.82} & \textbf{15875.72} & 0.839\\\hline
  \end{tabular}}
 \end{center}
\end{table*}

\begin{table*}[h]
	\caption{\blue{Evaluations of portfolio policy network with different feature extractors  in terms of APV, SR($\small{\%}$), CR and TO}}
 \label{table3}
 \vspace{-0.1in}
 \begin{center}
 \scalebox{0.78}{
 \begin{small}
	\begin{tabular}{|c|c|c|c|c|c|c|c|c|c|c|c|c|c|c|c|c|}\hline
  \multirow{2}{*}{Module} & \multicolumn{4}{c|}{Crypto-A}&\multicolumn{4}{c|}{Crypto-B}&\multicolumn{4}{c|}{Crypto-C}&\multicolumn{4}{c|}{Crypto-D}\cr \cline{2-17}
	   & APV & SR & CR & \blue{TO} & APV & SR & CR & \blue{TO} & APV & SR& CR &  \blue{TO} & APV & SR& CR   & \blue{TO}  \\\hline
      PPN-LSTM &  14.48  &  5.62 & 38.19   &0.459&      3550.32 & 13.75 &13297.32               &0.766& 2.85 & 3.99 & 6.69                   & 0.375& 159.54  &  15.16 & 6319.84         &0.786               \\\hline
      
      PPN-TCB &12.76 &  5.40  &26.52       &0.317&      3178.42 & 13.63 &  11011.87                 &0.753&         2.01 & 3.32 & 4.66             &0.211& 102.85  &  14.09 & 2972.63         &0.825        \\\hline
      
      PPN-TCCB & 16.51 &  6.01 & 35.89      &0.265&     4181.17 & 13.85 & 15798.05                 &0.642&       3.29 & 4.53 & 12.38                &0.560& 171.82  &  14.97 & 3945.99              &0.827       \\\hline
      
      PPN-TCB-LSTM & 18.62 & 6.28 & 39.87   &0.300&      4485.89 &14.18 & 15232.31                  &0.829&      3.49 & 4.48 &10.96                    &0.389& 179.43  & 15.18& 9150.33                 &0.788     \\\hline
      
      PPN-TCCB-LSTM &  21.03  &   6.12 & 52.75  &0.362&   5575.25 &  14.46 & 21353.23                  &0.728&     3.69 & 4.72 &   10.50                  &0.534& 224.41  &  15.99 &  8522.43         &0.881             \\\hline
      
      PPN-I & 25.76&  6.75   &57.05     &0.316&       7549.35 & 14.74 &  28915.43              &0.796&          3.93 & 5.12 & 15.04           &0.331& 238.93  & 16.07 & 8803.95      &0.937     \\\hline
      
      PPN &   \textbf{32.04} & \textbf{6.85} &  \textbf{79.87}   &0.368& \textbf{9842.56} & \textbf{14.82}  &\textbf{37700.03}      &0.888&\textbf{4.81} & \textbf{5.89} & \textbf{16.11}         &0.304& \textbf{538.22} & \textbf{17.82} & \textbf{15875.72} &0.839   \\\hline
	\end{tabular}
 \end{small}}
 \end{center}
\end{table*}

\begin{table*}[h]
	\caption{\blue{Comparisons under different transaction cost rates on  Crypto-A  in terms of APV, SR($\small{\%}$), CR and TO}}\vspace{-0.1in}
 \label{table4}
 
 \begin{center}
 \scalebox{0.82}{
 \begin{small}
	\begin{tabular}{|c|c|c|c|c|c|c|c|c|c|c|c|c|c|c|c|c|c|c|}\hline

   \multirow{2}{*}{Algos}&  \multicolumn{4}{c|}{c=0.01$\%$}& \multicolumn{4}{c|}{c=0.05$\%$} & \multicolumn{4}{c|}{c=0.1$\%$} & \multicolumn{4}{c|}{c=0.25$\%$}  \cr
   \cline{2-17}
   & APV & \blue{SR} & \blue{CR} & TO& APV & \blue{SR} & \blue{CR} & TO& APV & \blue{SR} & \blue{CR}& TO& APV &  \blue{SR} & \blue{CR} & TO  \cr \hline
       EIIE & 871.18 &  14.28& 3244.26 & 1.232  & 254.73 &11.59&942.02  & 1.076 & 77.79 & 9.38&  270.15&   0.859    & 10.48 &5.27&21.47& 0.471    \\\hline
       
       PPN-I& 1571.67 & 14.09&4728.58& 0.964 & 570.73 & 11.66&  1649.42&  0.779 & 219.30 & 10.13 & 663.81 & 0.668 &25.76 & 6.75 &57.05 & 0.316   \\\hline
       
       PPN & 3741.13  & 15.56 & 10593.34 & 1.018 & 754.57 &13.01 & 2821.25 & 0.731  & 242.27 & 10.52 & 864.17  & 0.658  &32.04 & 6.85 & 79.87 & 0.368   \\\hline  \hline 
    
    \multirow{2}{*}{Algos}&  \multicolumn{4}{c|}{c=0.5$\%$} & \multicolumn{4}{c|}{c=1$\%$}& \multicolumn{4}{c|}{c=2$\%$}& \multicolumn{4}{c|}{c=5$\%$}\cr
   \cline{2-17}
    & APV &\blue{SR} & \blue{CR} & TO& APV & \blue{SR} & \blue{CR} & TO& APV &  \blue{SR} & \blue{CR} & TO& APV & \blue{SR} & \blue{CR} & TO   \cr \hline
       EIIE & 1.95&2.45& 1.84 & 0.310  & 1.07 & 0.01& 1.00 & 0.247  & 0.81 &-0.54 & -0.37 & 0.021  & 0.28& -16.73 &-1.00 & 0.020    \\\hline
       
       PPN-I& 2.91 & 3.42&5.94&0.136 & 1.18 & 1.11 & 0.44  & 0.040 &0.96 & 0.33 & -0.15 & 0.013 & 0.99 & -2.92 & -0.99& 2e-7   \\\hline
       
       PPN & 2.92 & 3.11 & 4.42 & 0.239  & 1.61 & 2.01 & 2.16 &0.063  & 1.09 &  0.78 & 0.39 & 0.019  & 1.00 & 7.67 & 2.53 & 5e-8    \\\hline
	\end{tabular}
 \end{small}}
 \end{center}
\end{table*}
 
\begin{table*}[h]
	\caption{\blue{The performance of portfolio policy network under different $\gamma$ in terms of APV, SR($\small{\%}$), CR and TO}}
 \label{table5}\vspace{-0.1in}
 \begin{center}
 \scalebox{0.85}{
    \begin{small} 
    \begin{tabular}{|c|c|c|c|c|c|c|c|c|c|c|c|c|c|c|c|c|}\hline
  \multirow{2}{*}{$\gamma$} & \multicolumn{4}{c|}{Crypto-A}&\multicolumn{4}{c|}{Crypto-B}&\multicolumn{4}{c|}{Crypto-C}&\multicolumn{4}{c|}{Crypto-D}\cr \cline{2-17}
 & APV & \blue{SR}& \blue{CR} &TO& APV & \blue{SR}& \blue{CR}  &TO& APV & \blue{SR}& \blue{CR}  &TO& APV & \blue{SR}& \blue{CR}   &TO \\\hline
          10$^{-4}$ & 25.24 & 6.53 &66.46  &0.433   &      2080.69 &12.37&5668.71&  0.950    &        4.65 &5.88&19.34 &0.667    &        268.63 &16.89 & 1163.48 & 1.104 \\\hline
          
          10$^{-3}$ &\textbf{32.04} &\textbf{6.85} & \textbf{79.87}  &0.368 &   \textbf{9842.56}& \textbf{14.82} & \textbf{37700.03}  &0.888 &      \textbf{4.81} &\textbf{5.89}&\textbf{16.11}  &0.304 &     \textbf{538.22} & \textbf{17.82} & \textbf{2645.95}  &0.839  \\\hline
          
          10$^{-2}$ & 4.30 &  4.68 & 7.95 & 0.025   &      44.01  & 8.97 & 152.548 &0.161  &     1.21  & 1.62& 1.73 &0.008 &     1.72  & 4.02 & 9.33 &0.012 \\\hline
          
          10$^{-1}$ & 1.01  & 1.50&1.97 & 2e-08     &      1.65 & 2.81&2.50  &3e-03 &     1.00 &1.59& 0.29  &1e-7 &     1.00 & -0.08 & -4.35 &3e-7 \\\hline
	\end{tabular}
    \end{small}}
 \end{center}
\end{table*}	   
\begin{table*}[h]
	\caption{\blue{The performance of portfolio policy network under different $\lambda$ in terms of APV, STD($\small{\%}$), MDD($\small{\%}$) and TO}}
\vspace{-0.1in}
 \label{table6}
 \begin{center}
 \scalebox{0.87}{
 \begin{small}
    \begin{tabular}{|c|c|c|c|c|c|c|c|c|c|c|c|c|c|c|c|c|c|}\hline
   \multirow{2}{*}{$\lambda$} &\multicolumn{4}{c|}{Crypto-A}&\multicolumn{4}{c|}{Crypto-B}&\multicolumn{4}{c|}{Crypto-C}&\multicolumn{4}{c|}{Crypto-D}\cr \cline{2-17}
		& APV &STD& MDD & \blue{TO} &   APV &  STD& MDD & \blue{TO} & APV &STD  &MDD & \blue{TO} &  APV & STD& MDD & \blue{TO} \\\hline
          10$^{-4}$ & 32.04  & 2.16   & 38.86 &0.368&      9842.56  & 2.43   &  26.11  &0.888&      4.81  & 1.06   & 23.66 &0.304&       538.22  & 1.32   &  20.30 &0.839  \\\hline
          
          10$^{-3}$ & 25.56 & 1.99   & 37.40  &0.349&      8211.08  & 2.39   & 26.11 &0.645&      4.57   & 1.01   &  21.86 &0.341&      300.12  & 1.28   & 20.39 &0.934 \\\hline
          
          10$^{-2}$ & 25.38 & 1.95   & 37.26  &0.322&      4800.81  &  2.31   & \textbf{26.10} &0.779&      2.42  &  1.00   & 21.60  &0.334&       264.79  &1.27    &18.43 &0.789  \\\hline
          
          10$^{-1}$ & 9.81 & \textbf{1.85}    & \textbf{31.64} &0.543&     3353.55  & \textbf{2.30}   & \textbf{26.10} &0.682&      2.39  &\textbf{0.97}   & \textbf{20.12} &0.284&    195.75    &\textbf{1.17}   & \textbf{16.54} &0.931\\
    \hline
	\end{tabular}
 \end{small}}
 \end{center}
\end{table*}